\documentclass[acmlarge]{acmart}
\AtBeginDocument{%
  \providecommand\BibTeX{{%
    \normalfont B\kern-0.5em{\scshape i\kern-0.25em b}\kern-0.8em\TeX}}}

\setcopyright{acmcopyright}
\copyrightyear{xxxx}
\acmYear{xxxx}
\acmDOI{xx.xxxx/xxxxxxx.xxxxxxx}

\acmJournal{JACM}
\acmVolume{37}
\acmNumber{4}
\acmArticle{111}
\acmMonth{8}

\usepackage{booktabs}
\usepackage{multirow}
\usepackage{subfigure}
\usepackage{lineno}
\usepackage{algorithm}  
\usepackage{algpseudocode}  
\usepackage{amsmath}  
\usepackage{enumerate}
\usepackage[justification=centering]{caption}

\usepackage{footnote}
\usepackage{threeparttable}
\begin{document}

\title{Semi-Supervised Heterogeneous Graph Learning with Multi-level Data Augmentation}

\author{YING, CHEN}
\authornote{These authors contributed equally to this research.}
\email{kankan.cy@mybank.cn}
\author{SIWEI, QIANG}
\authornotemark[1]
\email{boyue.qsw@mybank.cn}
\affiliation{%
  \institution{MYbank, Ant Group}
  \city{Hangzhou}
  \state{Zhejiang}
  \country{China}
}

\author{MINGMING, HA}
\authornotemark[1]
\email{hamingming_0705@foxmail.com}
\affiliation{%
  \institution{School of Automation and Electrical Engineering, University of Science and Technology Beijing; MYbank, Ant Group}
  \city{Beijing}
  \country{China}
}

\author{XIAOLEI, LIU}
\authornotemark[1]
\email{liuxiaolei.lxl@mybank.cn}
\affiliation{%
  \institution{MYbank, Ant Group}
  \city{Beijing}
  \country{China}
}

\author{SHAOSHUAI, LI}
\email{lishaoshuai.lss@mybank.cn}
\author{JIABI, TONG}
\email{jiabi.tjb@mybank.cn}
\author{LINGFENG, YUAN}
\email{ryan.ylf@mybank.cn}
\affiliation{%
  \institution{MYbank, Ant Group}
  \city{beijing}
  \country{China}}
 
\author{XIAOBO, GUO}
\authornote{Corresponding author.}
\email{jefflittleguo.gxb@antgroup.com}
\affiliation{%
  \institution{Institute of Information Science, Beijing Jiaotong University; MYbank, Ant Group}
  \city{Beijing}
  \state{Beijing Shi}
  \country{China}}
  
\author{ZHENFENG, ZHU}
\email{zhfzhu@bjtu.edu.cn}
\affiliation{%
  \institution{Institute of Information Science, Beijing Jiaotong University}
  \city{Beijing}
  \state{Beijing Shi}
  \country{China}}

\renewcommand{\shortauthors}{Chen, et al.}

\begin{abstract}
In recent years, semi-supervised graph learning with data augmentation (DA) is currently the most commonly used and best-performing method to enhance model robustness in sparse scenarios with few labeled samples. However, most of existing DA methods are based on homogeneous graph while none are specific for heterogeneous graph. Differing from homogeneous graph, DA in heterogeneous graph has greater challenges: heterogeneity of information requires DA strategies to effectively handle heterogeneous relations, which considers the information contribution of different types of neighbors and edges to the target nodes. Furthermore, over-squashing of information is caused by the negative curvature that formed by the non-uniformity distribution and strong clustering in complex graph. To address these challenges, this paper presents a novel method named Semi-Supervised Heterogeneous Graph Learning with Multi-level Data Augmentation (HG-MDA). For the problem of heterogeneity of information in DA, node and topology augmentation strategies are proposed for the characteristics of heterogeneous graph. And meta-relation-based attention is applied as one of the indexes for selecting augmented nodes and edges. For the problem of over-squashing of information, triangle based edge adding and removing are designed to alleviate the negative curvature and bring the gain of topology. Finally, the loss function consists of the cross-entropy loss for labeled data and the consistency regularization for unlabeled data. In order to effectively fuse the prediction results of various DA strategies, the sharpening is used. Existing experiments on public datasets, i.e., ACM, DBLP, OGB, and industry dataset MB show that HG-MDA outperforms current SOTA models. Additionly, HG-MDA is applied to user identification in internet finance scenarios, helping the business to add 30\% key users, and increase loans and balances by 3.6\%, 11.1\%, and 9.8\%.
\end{abstract}



\received{20 February 2007}
\received[revised]{12 March 2009}
\received[accepted]{5 June 2009}

\maketitle

\section{Introduction}
In the real world, labeled data is limited, and most of the data are unlabeled \cite{miyato2018virtual}. Because relying on manual data labeling takes expensive time and cost. However, the traditional graph neural network only uses labeled data to supervise the loss minimization in training, and the unlabeled data does not participate. With this operation, the model is prone to over-fitting and poor generalization \cite{zhu2019robust}. Therefore, how to make full use of unlabeled data and labeled data for model training in graph neural networks is a problem that needs to be solved. As a promising strategy to leverage both limited labeled data and abundant unlabeled data, semi-supervised graph learning is proposed that supervised learning is performed on labeled samples, while consistent learning is conducted on unlabeled samples to learn the representations of latent key semantics in perturbation through DA \cite{ding2018semi,kipf2016semi,zhu2003semi}.

\begin{figure}
\centering
\includegraphics[width=0.8\textwidth]{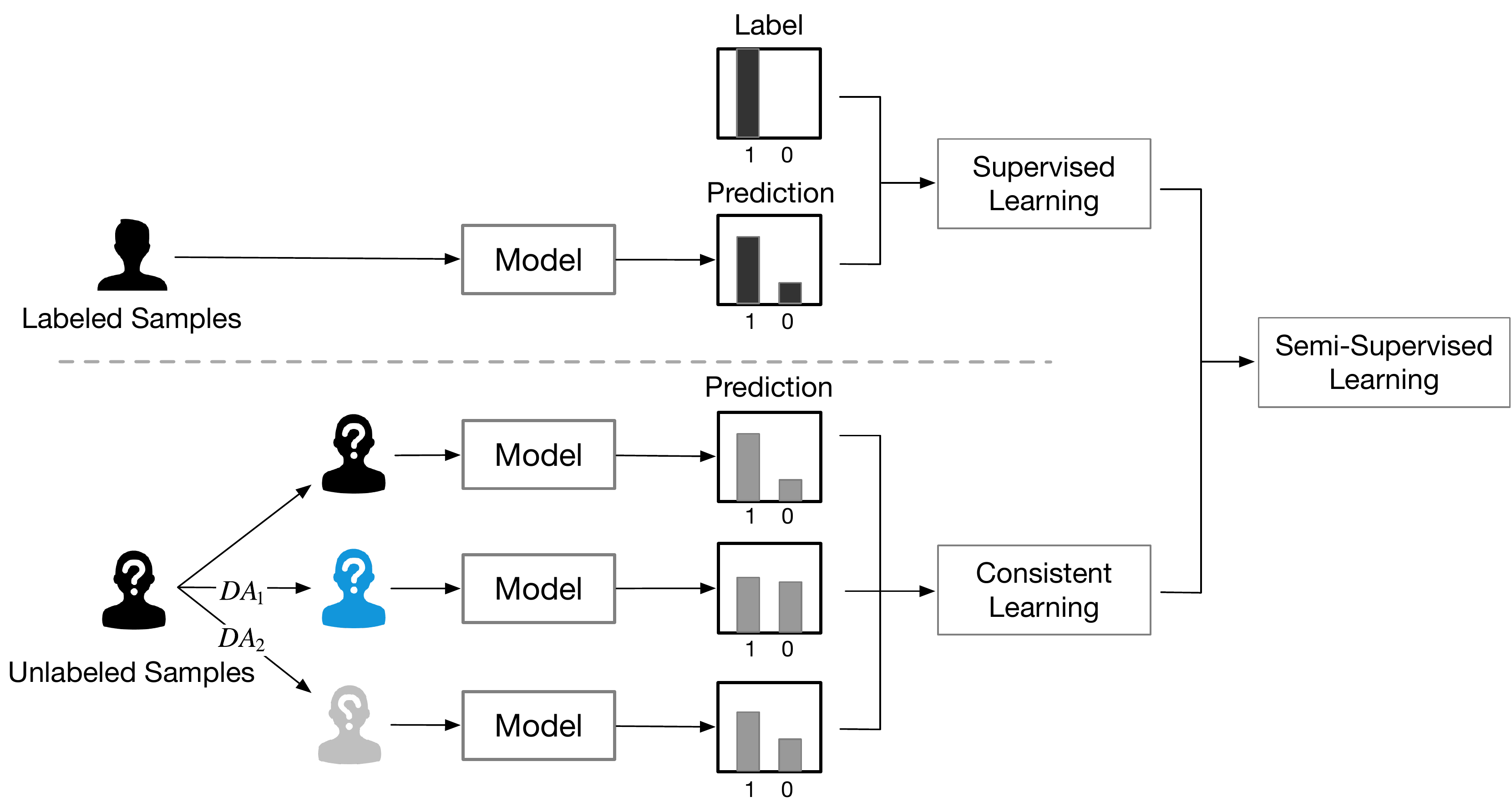}
\caption{\label{fig:intro}An Instance of Semi-Supervised Learning with DA. 
Semi-supervised learning with DA mainly includes two parts: supervised learning and consistent learning.}
\end{figure}

As mentioned by MixMatch \cite{2019MixMatch} and UDA \cite{2019Unsupervised}, forcing the model to be consistent through DA is currently the most commonly used and best-performing semi-supervised learning method. DA augments training data similar to real data through some data transformation methods. Its purpose is to generate diverse and realistic perturbations for unlabeled data, and to force the model to be consistent with respect to these perturbations, thereby enhancing the robustness of the model \cite{2019Unsupervised}. That is to say, a good model should be able to adapt to various small disturbances that do not change the nature of the data. Besides, DA can be regarded as a process of noise reduction and completion, reducing the noise or increasing the potentially important information in the graph. Referring to the instance in Figure \ref{fig:intro}, semi-supervised learning with DA mainly includes two parts: supervised learning and consistent learning. The supervised learning of labeled data makes the model classification more accurate, while consistency learning enables the model to extract valid signals from unlabeled data by applying DAs and constraining the outputs of unlabeled data before and after DAs to be as consistent as possible. Finally, the loss function is jointly optimized to transfer the label information from labeled data to unlabeled data, so that the model can extract effective information and learn diversity to improve the robustness of the model. Existing DA methods mainly involves node-level and topology-level augmentation strategies \cite{2021An, 2019sl}. Node augmentation strategies include the feature loss, the feature masking, and the feature exchange. Topology augmentation strategies include node removing, edge adding, and edge removing. It is considered that combining different levels of data augmentation can lead to more benefits on model learning \cite{2020Graph-GCL,2021An}. In recent research, GRAND \cite{feng2020graph} and NodeAug \cite{2020NodeAug} also tried to apply DA to homogeneous graph neural network and achieved great success. It was proved by experiments that DA played a crucial role in improving performance. At present, the DA technique for graph has not been fully explored yet. There are two problems here: 1) Most DA strategies are proposed for homogeneous graph, but not for the characteristics of heterogeneous graph. It is difficult to directly capture the important information derived from different types of nodes and edges. 2) Most DA strategies do not consider the effect on the network structure, which may further deteriorate the information over-squashing problem.

Heterogeneous information network, also referred to as heterogeneous graph for short, is more common in practical application scenarios, containing various types of entities and their various interaction relationships, and meanwhile, entities also contain the rich attribute information \cite{2016survey}. For example, an academic network contains different types of nodes with different attributes such as authors, papers, and topics, as well as different types of edges such as authors citing papers, papers belonging to topics. The heterogeneous graph is more generally applicable in real-world scenarios. From the previous analysis, it can be known that the heterogeneous graph has more complex properties than the homogeneous graph \cite{2019HetGNN,wang2022survey}. The representation learning of a node is affected by various aspects of information, such as the neighbors of the node, the features of the node, the sub-graph or paths where the node is located \cite{2016survey,hu2019cash}. In the process of information aggregation, nodes have different information preferences for different types of neighbors and edges. Hence, the challenges of DA for heterogeneous graphs are:

\begin{enumerate}[(1)]
\item Heterogeneity of information: Most DA strategies are proposed for homogeneous graph, but there are few studies dedicated to heterogeneous graph information networks. It is difficult to capture the important information of different types of nodes and edges when transplanting the general DA strategies directly to heterogeneous graph. Therefore, it is necessary to propose corresponding DA strategies for the characteristics of heterogeneous graphs. 
\item Over-squashing of information: The non-uniformity distribution of degrees of nodes in complex networks causes negative curvature of graph networks, which are mainly characterized by power-law distribution and strong clustering properties \cite{2003Hierarchical}. Negative curvature can cause the bottlenecks of graph information and lead to the phenomenon of over-squashing of information, which greatly reduces the learning performance of the model \cite{topping2021understanding}. Current DA strategies do not consider the impact on the network structure, which may further exacerbate the problem of excessive information compression.
\end{enumerate}

To address the aforementioned challenges, a novel semi-supervised Heterogeneous Graph learning with Multi-level Data Augmentation is proposed, namely HG-MDA. HG-MDA is a SSGL-DA framework widely applicable to heterogeneous graphs that employs multi-level DA strategies to adequately learn the node attributes and structural information of labeled and unlabeled target nodes. Specifically, In HG-MDA, the meta-relation-based attention is applied to adaptively learn the heterogeneous information strength. Based on this, the node-level and topology-level DA strategies, and corresponding objective functions are designed, respectively. Node augmentation introduces the feature similarity to measure the feature distribution between different types of nodes, so as to select the augmented node to exchange features with the target node. Differing from other edge adding/removing strategies, triangular augmentation takes into account the different contributions of the different types of neighbor nodes and edges to node information aggregation. The neighbor nodes and the corresponding edges that transmit more information to the target node are selected for edge adding, and the edges with small information are removed, so as to introduce more effective information gain for heterogeneous graphs. In order to integrate different DA methods, a sharpening operation is added to calculate the consistency regularization loss, which makes the difference in the probability of different DA methods more obvious in each category. Overall, the major contributions of this paper includes three aspects:

\begin{enumerate}[(1)]
\item A novel framework HG-MDA for heterogeneous graphs is proposed. Node augmentation and triangle augmentation enable the model to effectively learn the node attributes and structural information of labeled and unlabeled target nodes. Moreover, the combination of multi-level DA and the designed meta-relation-based attention enables the preferences of different types of nodes and edges to be captured during DA.
\item This paper is the first exploratory work on triangle-based data augmentation. Triangle augmentation can change the number of triangles and open triangles in the graph. The present triangular augmentation can smooth the degree distribution of nodes, adjust the graph curvature and optimize network information transmission, so as to solve the problem of information compression and make representation more accurate.
\item Experiments are performed on node classification and node clustering tasks in public datasets such as ACM, DBLP, OGB, and real-world industry datasets MB. The experiment results show that the proposed approach outperforms the current SOTA model.
\end{enumerate}

\section{Related Works}
In this section, some existing related works are reviewed, including semi-supervised learning and graph contrastive learning. At last, the existing problems in semi-supervised learning and DA are discussed and the comparisons between the proposed model and related work are summarized.

\subsection{Semi-Supervised Graph Learning}

Semi-Supervised Graph Learning (SSGL) is a semi-supervised learning for graph data, which can effectively combine labeled data and unlabeled data for learning. It is suitable for scenarios where there is a small amount of labeled data and a large amount of unlabeled data, and the distribution of unlabeled data and labeled data is the same. Although the model can achieve good performance with only labeled data, making full use of unlabeled data through SSGL can reduce the risk of model over-fitting and further improve model performance \cite{zhang2022semi,yang2021survey}. SSGL can be divided into two types: graph neural network and semi-supervised graph learning with data augmentation.

\subsubsection{Graph Neural Network}

Graph Neural Networks (GNN) are the most primitive semi-supervised deep learning methods for graph data \cite{wang2022survey}. GNNs can learn the information of unlabeled nodes because the graph structure reflects the mutual links between unlabeled nodes and labeled nodes, and the representation of nodes are updated by sharing parameters, so that the information of unlabeled nodes in the process of aggregation can be passed to labeled node. For heterogeneous graphs, Wang et al. proposed Heterogeneous Graph Attention Network (HAN) \cite{2019HAN}, which for the first time extended GNNs to heterogeneous graphs and designed a hierarchical attention mechanism that can capture both structural and semantic information. Subsequently, a series of Heterogeneous Graph Neural Networks (HGNNs) were proposed \cite{2020MAGNN,hu2019cash,2019Heterogeneous,2020HGT}. Some HGNNs like HAT are mainly based on attention to aggregate different semantic content between different types of nodes and different information from different paths. MAGNN \cite{2020MAGNN} designs three main components: a node content transformation component that unifies semantics and dimensions, a intra-metapath aggregation component that merges the embeddings of intermediate nodes, and a inter-metapath aggregation component that merges information from multiple meta-paths. However, the choice of metapath is still an open issue, which relies on either domain knowledge or label information. So HGT \cite{2020HGT} abandons mate-paths and introduces the adaptive attention that depends on the types of source nodes, target nodes and edge types instead. The meta-paths are not artificially designed and the information of different types of high-order neighbors can be merged. In addition, there are some HGNNs that focus on other problems. GTN focuses on automatically identifying useful meta-paths and higher-order links during learning node embeddings \cite{2019GTN}. RGCN uses multiple weight matrices to embed nodes into different relational spaces, thus capturing the heterogeneity of graphs \cite{2018RGCN}.

The semi-supervised learning based on GNNs/HGNNs has limitations. Its semi-supervision is mainly reflected in the graph structure composed of unlabeled nodes and labeled nodes, but the objective function only uses labeled nodes for supervision, resulting that the model is prone to bias when there is less labeled data and more unlabeled data. Therefore, semi-supervised graph learning with data augmentation has gradually attracted attention in recent years.

\subsubsection{Semi-Supervised Graph Learning with Data Augmentation}

The core ideas of Semi-Supervised Graph Learning with Data Augmentation (SSGL-DA) is similar to the Semi-Supervised Learning for structured data \cite{2013pseudo,2015ladder,2019MixMatch,2019Unsupervised}. First, the DA strategies are used to introduce data perturbation so that the model can extract effective signals from the data. Second, the objective function consists of a supervised loss for labeled data and a regularization loss for unlabeled data. Regularization ensures that multiple predictions of the model for unsupervised data can be stabilized on the same category, indicating that the model has a certain discriminative ability in the learning of unlabeled data \cite{2004entropy}. SSGL-DA implicitly enhances the generalization ability of GNNs \cite{feng2020graph}.

In GRAND \cite{feng2020graph}, a random propagation strategy is designed to add noise to the original graph, which interferes with the feature propagation on the graph, making each node insensitive to a specific neighborhood and enhancing the robustness of GRAND. In order to make full use of the large amount of unlabeled data in graph learning, GRAND draws on the idea of MixMatch/UDA that consistency regularization is used to optimize the prediction consistency of unlabeled nodes in different DAs \cite{2019MixMatch,2019Unsupervised}. Compared with the random augmentation strategy and general regularization method of GRAND, NASA \cite{2022NASA} proposes to replace adjacent nodes with far-neighbor nodes to interfere with nodes in DA and a neighborhood-constrained regularization method that supervises representations of neighbor nodes to be consistent with the representation of the target node in regularization. NodeAug \cite{2020NodeAug} mainly focuses on the design of DA and proposes three DA approaches, namely node attributes replacing, edge adding and edge removing. The importance of node features is measured by the weight matrix and the importance of edges is measured by the degree of the node. Besides, NodeAug believes that there is mutual influence between different DAs in a graph, so it proposes a node parallel augmentation scheme to ensure that the DAs of different nodes will not affect each other.

But there are also some problems. The augmentation strategies of GRAND and NASA are stochastic, which does not guarantee that the perturbations are useful to the model. Although NodeAug uses the weight matrix and the degree of the node to measure the importance of node features and edges, it is not comprehensive enough. The weight matrix is shared by the whole graph, reflecting the global importance of the node, but not the importance of the node to the specific target node. And measuring the importance of an edge by the degree of a node also does not take into account the relation between the node and the target node and the augmented network structure. Therefore, the proposed method in this paper adopts the meta-relation-based attention to measure the importance of node features and edges. The attention represents the importance of the target node under the relation of each type of node and edge, which better reflects the influence of heterogeneity. In addition, edge adding and removing based on the triangle structure are proposed, so as to consider the connection between nodes and optimize the entire network structure.

\subsection{Graph Contrastive Learning}

The DA strategies in SSGL-DA are similar to Graph Contrastive Learning (GCL), so in-depth research on GCL has been conducted to further expand ideas. An Empirical Study \cite{2021An} provides a comprehensive introduction to several key parts of GCL involving DA strategies, contrast modes, contrast targets and negative mining techniques. The effects of different GCL components on the model were investigated through extensive controlled experiments. Three of the conclusions obtained in this paper have brought us inspiration. (1) The biggest benefit of the GCL algorithm is the topology augmentation with sparse views (such as edge removing). (2) Two-layer DA at the node-level and the topology-level can further improve the performance. (3) The comparison mode of the same dimension should be selected according to the granularity of downstream tasks, such as node-node or graph-graph comparison. The paper also develops an easy-to-use library PyGCL, with modular components, standardized evaluation and experiment management. In the Section \ref{experiments}, GCA \cite{2021GCA}, GRACE \cite{2020Deep}, BGRL \cite{2021Bootstrapped}, and G-BT \cite{2021GBT} provided by PyGCL are used as comparison methods. Another paper \cite{2020Graph-GCL} also systematically studied the effect of different graph augmentations. The conclusions it draws also help us: (1) DA is crucial in GCL. The original model combined with appropriate DA strategies can improve the model prediction performance. (2) Combining different DAs can bring more gains. (3) Edge perturbations are beneficial for social network prediction.

Referring to these experiences, the proposed method applies data augmentation at the node level and topology level, including three strategies: feature exchange, edge adding, and edge removing. Our experimental performance is also in line with the above conclusions, proving that DA is the key to improving the performance of the model.

\section{Our Approach}

This section mainly introduces semi-supervised heterogeneous graph learning with multi-level DA. First, the relevant basic definitions are introduced. Secondly, it overviews the overall structure of the model, and then introduces each part of the model in detail.

\subsection{Problem Definition}

Heterogeneous information network for SSL is defined as $G=(\mathcal{V}, \mathcal{E}, \mathcal{X})$, in which $\mathcal{V}$ is a node set, $\mathcal{E}$ is a edge set and $\mathcal{X}$ is a feature set corresponding to $\mathcal{V}$. For the target node set $T\in \mathcal{V}$, there are labeled nodes $T_l$ and unlabeled nodes $T_u$ in $T$, and the number $\vert{T}_l\vert$ is much smaller than $\vert{T}_u\vert$. Besides, each heterogeneous graph is associated with a node type mapping function $\mu_1: \mathcal{V} \to \mathcal{A}$, $\mu_2: \mathcal{X} \to \mathcal{A}$ and a edge type mapping function $\mu_3:\mathcal{E} \to \mathcal{R}$ as well. $\mathcal{A}$ and $\mathcal{R}$ denote the sets of predefined node types and edge types, having constrains $\vert\mathcal{A}\vert + \vert\mathcal{R}\vert > 2$.

\begin{figure}[htbp]
\centering
\subfigure[\label{fig:triangle}The Triangle Structure.]{
\begin{minipage}[t]{0.4\linewidth}
\centering
\includegraphics[width=0.3\textwidth]{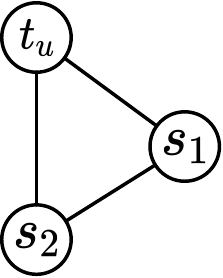}
\end{minipage}%
}%
\subfigure[\label{fig:v-structure}The Open Triangle Structure.]{
\begin{minipage}[t]{0.4\linewidth}
\centering
\includegraphics[width=0.3\textwidth]{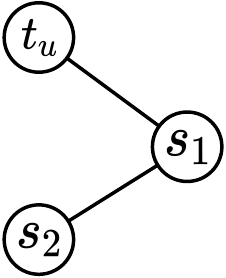}
\end{minipage}%
}%
\centering
\caption{\label{fig:structure}The Definition of Triangle Structure and Open Triangle Structure.}
\end{figure}

In order to solve the problem of heterogeneity of information in DA and over-squashing of information, node augmentation and triangle augmentation are proposed. Node augmentation is defined as feature exchange between an unlabeled node $t_u \in T_u$ and its neighbor node set $\mathcal{S}={s_1,s_2,...,s_n}$, where $X_u$ is the feature set corresponding to $t_u$ and $\mathcal{X}_\mathcal{S}=X_{s_1},X_{s_2},...,X_{s_K}$ is corresponding to $\mathcal{S}$. According to the selection metrics, the most relevant $K$ neighbor nodes $S_K={s_1,s_2,...,s_K}$ are selected. And then the features $X_\mathcal{S}$ are averaged to get the final aggregated feature $X_exchange$ to exchange with the target node $X_u$. Triangle augmentation is the operation of adding or removing edges based on the triangle structure related to unlabeled nodes. The definitions of triangles and open triangles are given here. For unlabeled node $t_u$, let the 2-hop neighbor node set of $t_u$ denote $\mathcal{S}={s_1,s_2}$. The edges $(t_u, s_1), (s_1, s_2), (t_u, s_2)$ form a triangle, also known as a closed triangle. There is an edge connection between any two nodes, as shown in Figure \ref{fig:triangle}. On the other hand, the edges $(t_u, s_1), (s_1, s_2)$ form an open triangle, as shown in Figure \ref{fig:v-structure}. For the target node $t_u$, as node $s_1$ is the 1-hop node and node $s_2$ is the 2-hop node, there is no direct connection between the target node $t_u$ and the 2-hop node $s_2$.

\subsection{The Overall Architecture}

\begin{figure}
\centering
\includegraphics[width=1\textwidth]{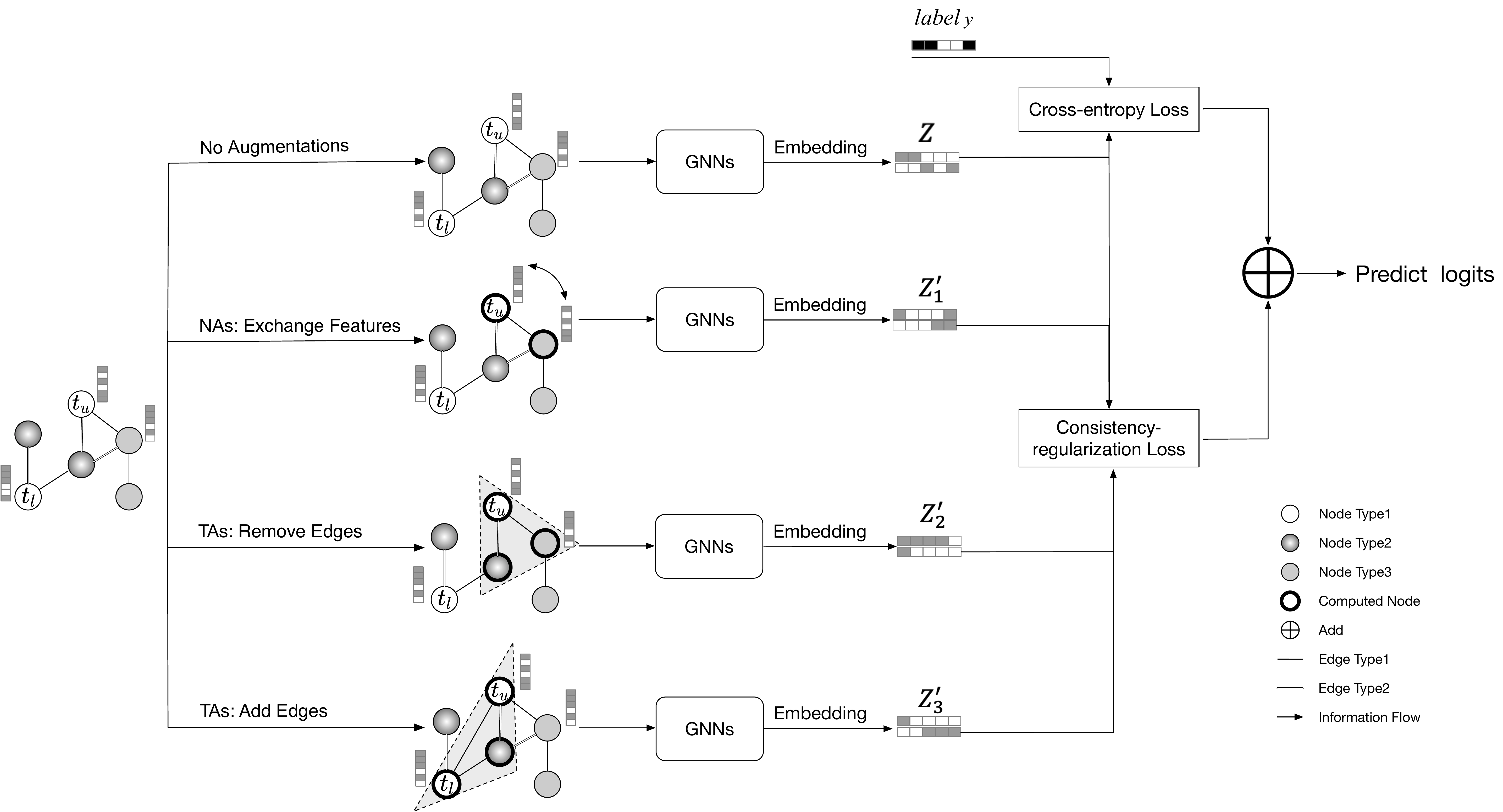}
\caption{\label{fig:overall}The Overall Architecture of HG-MDA.
\textsl{"NAs" means Node Augmentations and "TAs" means Triangle Augmentations. Given a sampled heterogeneous sub-graph with $t_u$ as a unlabeled target node, HG-MDA obtains different embedding vectors of $t_u$ through node augmentation and triangle augmentation respectively. For labeled nodes $t_l$, supervise them with ground truth by cross entropy loss. And for unlabeled nodes $t_u$, supervise different embedding before and after augmentation by consistency regularization loss.}}
\end{figure}

In our approach, Semi-Supervised Heterogeneous Graph Learning with Multi-level DA(HG-MDA) is proposed in which unlabeled data and labeled data are used for heterogeneous graph model training. Figure \ref{fig:overall} shows the overall architecture of HG-MDA. Such process can be decomposed into four components: Meta-relation-based Attention, Multi-Level DA that includes Node Augmentation and Triangle Augmentation, and Loss Function. Given a sampled heterogeneous sub-graph that contains three types of nodes and two types of edges, the attention under each triple of <neighbor type, edge type, target type> is learned based on the meta relation according to the characteristics of heterogeneous graph. On the one hand, it is used for the aggregation of heterogeneous information. On the other hand, it is used as a selection index to measure the importance of nodes and edges in the next node augmentation and triangle augmentation. And then node augmentation and triangle augmentation are designed based on the attention. In the node augmentation, the similarity of the features of neighbor nodes and the target nodes is proposed, combining the meta-relation-based attention to select the neighbor nodes. It ensures that the selected node is the strongest information and the most similar features to exchange features with the target nodes. In the triangle augmentation, the Adamic-Adar index is introduced to choose the 2-hop nodes for edge adding to form triangles. And the meta-relation-based attention is applied to choose the type of the added/removed edges. Moreover, a loss function is designed to supervise the training of labeled data and unlabeled data at the same time. To effectively integrate different DA strategies, the results of multiple data augmentations are processed through sharpening in the loss function of unlabeled data. The goal of HG-MDA is to introduce unlabeled data to enhance generalization performance through DA. 

\subsection{Meta-relation-based Attention}

In heterogeneous graph learning, different types of nodes and edges have different information strength to the target node, so it is not only necessary to learn the information strength of different paths to the target node in message passing, but also in DA. Inspired by the attention design of Transformer, multi-head attention based on meta relation is taken into DA to learn the information strength of the aggregated message under different paths to the target node. A meta relation in heterogeneous network is defined as a triple consisting of $<\mathcal T(s), e, \mathcal T(t)>$. Let $s$ be a source node with type $\mathcal T(s)$, $t$ be a target node with type $\mathcal T(t)$ and $e$ be the edge type between $s$ and $t$.

\begin{figure}
\centering
\includegraphics[width=1\textwidth]{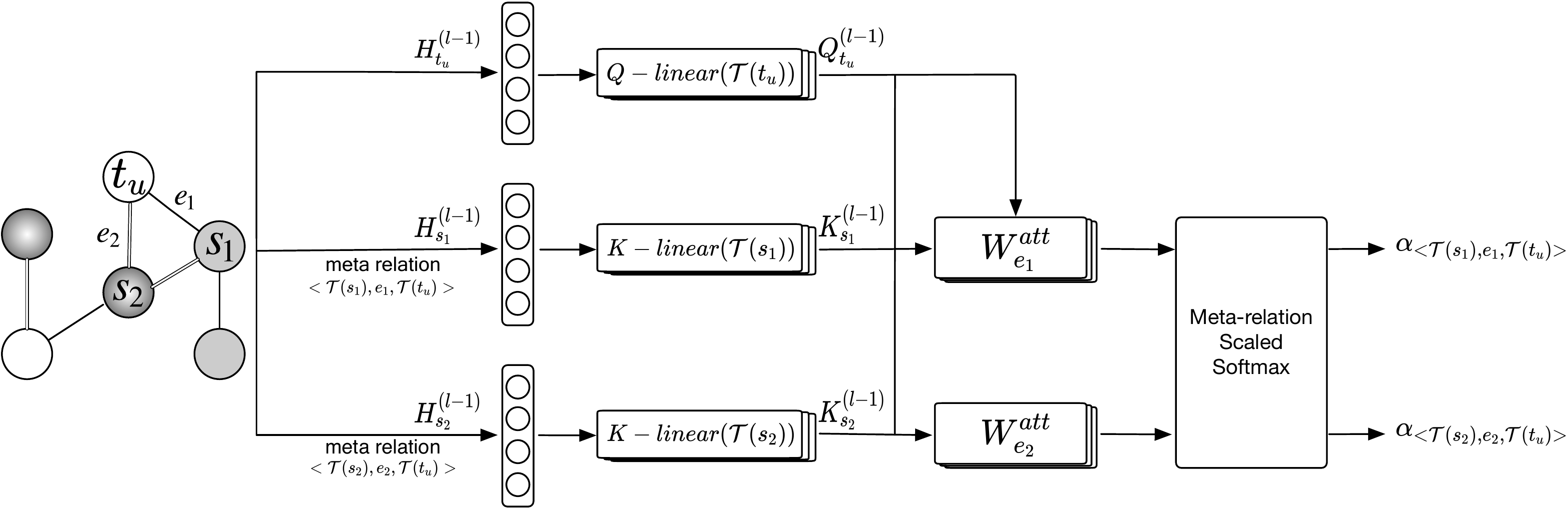}
\caption{\label{fig:attention}The Calculation Process of Meta-relation-based Attention.
\textsl{Let $t$ be the target node with type $\mathcal T(t)$, $s_i$ be the source node with type $\mathcal T(s_i)$, $i=1, 2, \ldots, I$ denotes the number of source node types. Attention takes meta relation of $<\mathcal T(s_i), e_k, \mathcal T(t)>$ as input to learn information strength passed by $s_i$ through the edge type $e_k$, $k=1, 2, \ldots, K$ denotes the number of edge types.}}
\end{figure}

Triples with the same meta relation share the same attention. Figure \ref{fig:attention} shows the calculation process of meta-relation-based attention.

Different types of nodes have different feature dimensions. First, a projection $W_{\mathcal T(*)}\in{\mathcal R^{{d_{\mathcal T(t)}}\times d}}$ is used to project the $d_{\mathcal T(*)}$-dimension node vectors of type $\mathcal T(*)$ to the same dimension $d$, so that different types of nodes are comparable. Suppose $X_{\mathcal T(*)}$ is a node vector of type $\mathcal T(*)$ 
\begin{equation}
H^0 = \sigma(W_{\mathcal T(*)}X_{\mathcal T(*)}).
\end{equation}

Let $H_t^{(l-1)}$ and $H_{s_i}^{(l-1)}$ be the $d$-dimension vectors of $t$ and $s_i$ in the $l$-th layer. For the target node, $H_t^{(l-1)}$ is projected with the linear projection $Q-linear(\mathcal T(t))$ to obtain $\frac{d}{H}$-dimensional vectors $Q_t^{(l-1)}$. For the source node, $H_{s_i}^{(l-1)}$ is projected with the linear projection $K-linear(\mathcal T(s_i))$ to obtain $\frac{d}{H}$-dimensional vectors $K_{s_i}^{(l-1)}$ ($H$ is the number of multi-head attention, and $\frac{d}{H}$ is the vector dimension per attention head). The information strength of $K_{s_i}^{(l-1)}$ to $Q_{t_u}^{(l-1)}$ for $h$-th attention head is calculated by 

\begin{equation}
attention^h_{<\mathcal T(s_i), e_k, \mathcal T(t)>} = (K_{s_i}^{(l-1)}W_{e_i}^{att}Q_{t_u}^{(l-1)})\frac{V_{<\mathcal T(s_i), e_k, \mathcal T(t)>}}{\sqrt{d}}.
\end{equation}

In a heterogeneous graph, there are multiple types of edges between nodes, which may have different dimensions. So a parameter matrix $W_{e_i}^{att}\in{\mathcal R^{\frac{d}{H}\times\frac{d}{H}}}$ for each edge type $e_i$ is needed here to capture different semantic relation between two nodes, even between the same node type pairs. Besides, considering that the number of edge types between each node type pairs is different, the vector $V_{<\mathcal T(s_i), e_k, \mathcal T(t)>}$ is used to unify the output of each meta relation $<\mathcal T(s_i), e_k, \mathcal T(t)>$ into the same dimensions, and $\frac{1}{\sqrt{d}}$ for numerical scaling.

The outputs of attention heads are inputted to $softmax$ to get the final attention weight of per meta relation $<\mathcal T(s_i), e_k, \mathcal T(t)>$ to the target node $t$. That is, for all neighbor node types $\mathcal T(s_i)$ and corresponding edge types $e_k$ of target node t, there are 

\begin{equation}
\alpha_{<\mathcal T(s_i), e_k, \mathcal T(t)>} = softmax(attention^h_{<\mathcal T(s_i), e_k, \mathcal T(t)>}).
\end{equation}

\subsection{Node Augmentation}

Node augmentation mainly transforms the attributes of the node itself to obtain training data that is similar to the real data distribution but introduces some disturbance. DA for feature exchange is considered in our approach. It can be seen at  that 1-hop nodes can provide richer and more direct information for the target node. So for the nodes to be augmented, we mainly consider the feature exchange with the 1-hop neighbor nodes.

In the process of feature exchange, meta-relation-based attention is used to measure the information strength that the features of different types of neighbor nodes with different edge types are finally transmitted to the target node. Feature similarity between neighbor nodes and target nodes is proposed and used together with meta-relation-based attention to select the neighbor nodes that can provide the strongest information and the most similar features to exchange features with the target nodes, thereby introducing perturbations on the features of the nodes.

\begin{figure}
\centering
\includegraphics[width=1\textwidth]{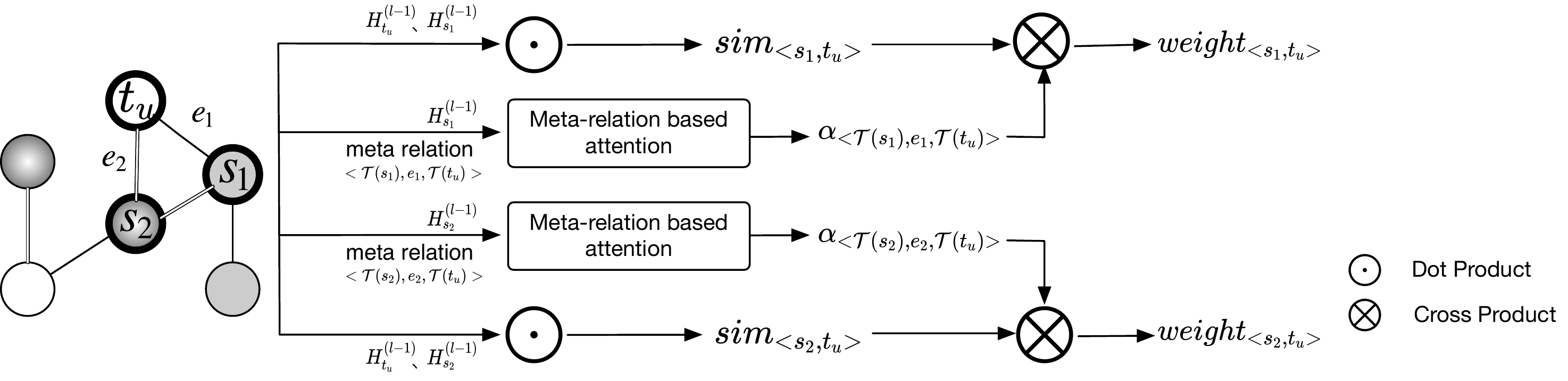}
\caption{\label{fig:feature_exchange}The Calculation Process of The Contribution in Feature Exchange.
\textsl{The neighbor nodes $H_{s_1}^{(l-1)}$ and $H_{s_2}^{(l-1)}$ are input into the attention mechanism to get the contribution $\alpha_{<\mathcal T(s_1), e_1, \mathcal T(t_u)>}^{(l-1)}$ and $\alpha_{<\mathcal T(s_2), e_2, \mathcal T(t_u)>}^{(l-1)}$. In addition, for the unlabeled target node $t_u$, the dot product $sim_{<s_1,t_u>}^{(l-1)}$ of the target node $H_{t_u}^{(l-1)}$ and $H_{s_1}^{(l-1)}$ 
multiply by $\alpha_{<\mathcal T(s_1), e_1, \mathcal T(t_u)>}^{(l-1)}$ to get the final features information strength $weight_{<s_1,t_u>}$, so does the dot product $sim_{<s_2,t_u>}^{(l-1)}$ of the target node $H_{t_u}^{(l-1)}$ and $H_{s_2}^{(l-1)}$.}}
\end{figure}

Figure \ref{fig:feature_exchange} shows the calculation process of the features information strength in feature exchange. For the unlabeled target node $t_u$ to be augmented, meta-relation-based attention is used to calculate the information strength $\alpha_{<\mathcal T(s_i), e_k, \mathcal T(t_u)>}$ of each meta relation $<\mathcal T(s_i), e_k, \mathcal T(t_u)>$ to the target node $t_u$ in the message aggregation of neighbor nodes. And then the dot product of the vector of the neighbor node and the target node as a similarity measure of features. The feature similarity is calculated by 

\begin{equation}
sim_{<s_i,t_u>} = H_{t_u} \cdot H_{s_i}.
\end{equation}

Finally, the feature similarity $sim_{<s_i,t_u>}$ and information strength $\alpha_{<\mathcal T(s_i), e_k, \mathcal T(t_u)>}$ multiply to obtain the final information strength of features. 

\begin{equation}
weight_{<s_i,t_u>} = \alpha_{<\mathcal T(s_i), e_k, \mathcal T(t_u)>}sim_{<s_i,t_u>}
\end{equation}

Finally, a information strength vector $W\in\mathcal R^n$ of all neighbor nodes is obtained, where $n$ is the number of neighbor nodes. The top $K$ neighbor nodes with the greatest contribution are selected to exchange features with the target node, and take the mean of the neighbor features as the feature of the target node. The aggregation of $K$ neighbor features is averaged over the columns of the matrix $\mathcal X_K$, which is given by

\begin{equation}
\mathcal X_{exchange} = mean(\mathcal X_K)
\end{equation}

We considers that the features of neighbor nodes with the largest $weight_{<s_i,t_u>}$ is most similar to the target node's and is the most important in the aggregation process, so that it can introduce a little disturbances after feature exchange. Based on the assumption of graph homology: adjacent nodes often have similar features, and the model can recover the exchanged feature information of the target node through the context of the remaining nodes \cite{2014deepwalk,feng2020graph}.

\subsection{Triangle Augmentation}

This subsection introduces specific schemes of triangle augmentation. Firstly, the implementation details of triangle based edge adding and removing are introduced, respectively. Then, the benefit of triangle augmentation is explained based on geometric analysis.

\subsubsection{Triangle based Edge Adding}

One of the ideas proposed in structure learning is that the original graph structure is not necessarily reliable \cite{2022Compact}. There is a common problem that the constructed graph is incomplete due to noise or data loss or incomplete information. Therefore, how to learn a suitable graph structure instead of relying on the original graph structure is a key issue.

In actual scenarios, the constructed graph may be incomplete due to data loss or indirect influence. For example, an author cites an formula in paper $A$, but the formula in paper $A$ actually refers to paper $B$. Although there is no direct connection between paper $B$ and the author, there is actually information transfer between paper $B$ and the author. Therefore, the purpose of edge adding is to restore the real triangular structure in graph.

\begin{figure}
\centering
\includegraphics[width=1\textwidth]{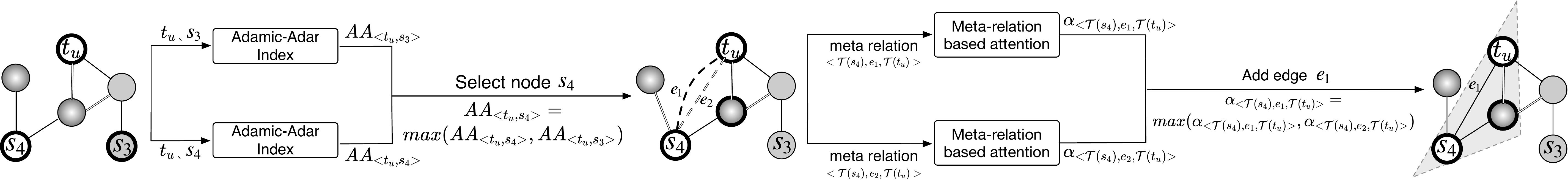}
\caption{\label{fig:edge_adding}The Process of Triangle based Edge Adding.
\textsl{For a unlabeled target node $t_u$, triangle based edge adding include three components: 1. Calculate the $AA$ Indexes of the target node and all 2-hop nodes of the open triangles to get the vector $V_A = (AA_{<t_u,s_1>},AA_{<t_u,s_2>},\cdots,AA_{<t_u,s_n>})$, $n$ is the number of the 2-hop nodes. 2. Sort $V_A$ in descending order and select the top $K$ nodes to add edges to the target node. 3. Select the types of added edges by meta-relation-based attention. $A \ni \alpha_{<\mathcal T(s_i), e_k, \mathcal T(t_u)>}$ is a vector including attention weights of all meta relations of 2-hop nodes to the target node. Sort $A$ in descending order and the top $K$ most important edge types are used as new edge types.}}
\end{figure}

Whether it is possible to add an edge between two nodes to form a triangular structure, in heterogeneous graph, it is necessary to consider how to select the second-order node that adds an edge to the target node and the type of the added edge. Figure \ref{fig:edge_adding} shows the process of triangle based edge adding.

First, we introduce how to select the node that adds an edge to the target node. In multiple open triangles, as shown on the left of Figure \ref{fig:edge_adding}, an importance measure is required to measure whether there is a relation between the target node and the second-order node. The most famous one in practical applications is the Adamic-Adar Index ($AA$ Index) \cite{2003Adamic}. Compared with the similarity indicators such as Common Neighbors, Jaccard's Coefficient, Preferred Attachment, and Cosine Similarity \cite{samanta2018link}, $AA$ index not only considers the number of common friends between nodes, but also considers the degree of the node itself. Nodes with a relatively large degree are penalized, reducing the influence of public nodes with very large neighborhoods, and increasing the attention of nodes that only share with a small number of nodes. $AA$ index is simple, but very effective, so $AA$ is used as a measure of relationship prediction between nodes in this paper.

In the open triangle, let $t$ be the target node, $s$ be the source node and $z$ is the common neighbor node between $t$ and $s$. $AA$ index is calculated by \begin{equation}
AA_{<t,s>} = \sum_{z\in\Gamma(t)\cap\Gamma(s)} {\frac{1}{log(\left|\Gamma(z)\right|)}}
\end{equation}

$\Gamma(z)$ denotes the number of $z$'s neighbors. The $AA$ index uses the degree of the intermediate node as a weight. The greater degree of the intermediate node, the smaller weight. The $AA$ index is used to measure the closeness of the source node to the target node. The closer a node is to the target node, the more likely the target node is to connect with it. Therefore, the top $K$ source nodes with largest $AA$ index are selected to add edges to the target node. $K$ denotes the augmentation ratio of source nodes that need to add edges.

Then, the types of the added edges need to be considered. If there is no or only one meta relation between the target node type and the source node type, the corresponding edge can be added directly. But if there are multiple types of meta relation between the two node types, the edge types need to be determined by the information strength $\alpha_{<\mathcal T(s_i), e_k, \mathcal T(t)>}$ from meta-relation-based attention. And then the top $K$ edge types with most important are used as new edge types. 

\subsubsection{Triangle based Edge Removing}

In fact, a triangle may also be formed due to noise. For example, in the triangle formed by APP-store-customer, a general APP is widely used by many people, so it will be connected with the store and also with customers. In addition, the customer's accidental purchase behavior causes a link between the customer and the store. But the customer has no relation with the store actually, so this link is noise data. Therefore, edge removing is the process of removing noise.

As mentioned above, 1-hop nodes and edges are closer to the target node and provide more information. If there is noise, the interference is also the largest. Therefore, edges between the 1-hop nodes and the target node are considered to remove. A target node may be in multiple triangles. Which edges are noise for the target node needs to be judged. 

\begin{figure}
\centering
\includegraphics[width=1\textwidth]{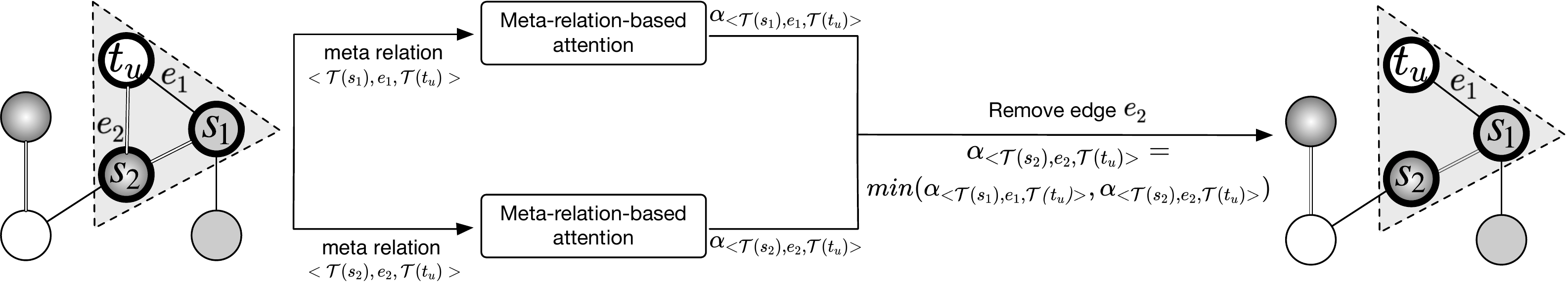}
\caption{\label{fig:edge_removing}The Process of Triangle based Edge Removing.
\textsl{For a unlabeled target node $t_u$, triangle based edge removing include two components: 1. Select the types of removed edges by meta-relation-based attention $A$. $A$ is a vector set of attention weights $\alpha_{<\mathcal T(s_i), e_k, \mathcal T(t_u)>}$ of all meta relation of 1-hop nodes to the target node. 2. Sort $A$ and the top $K$ edge types with less importance are removed.}}
\end{figure}

The process of triangle based edge removing is shown as Figure \ref{fig:edge_removing}. First, Triangles containing the target node in graph are found, judge the importance of the edges to the target node according to meta-relation-based attention. And then the top $K$ edges with less importance is selected to remove. $K$ denotes the augmentation ratio of source nodes that need to removing edges and is set the same as adding edges.

\subsubsection{\label{geometric}Geometric Analysis} 

Triangle-based augmentation can alleviate the problem of over-squashing of information and introduce new information gain for the graph. Before explaining the benefits of triangle augmentation, the definition of information compression and the relation between curvature and clustering coefficients need to be understand.

In some recent studies, it has been pointed out that in graph neural networks, the representation learning of nodes depends on the information of nodes connected to them remotely. Especially for a ``small world'' graph like a social network, if there is a long-term dependency between nodes and the degree distribution of neighbor nodes at different levels is uneven, it will cause the size of the nodes' receptive field to grow exponentially with the order, leading to over-squashing of information during the process of aggregating neighbor node information and learning fixed-size node embeddings \cite{alon2020bottleneck}.

\begin{figure}[htbp]
\centering
\subfigure[\label{fig:cur-1}Negative Curvature]{
\begin{minipage}[t]{0.4\linewidth}
\centering
\includegraphics[width=0.4\textwidth]{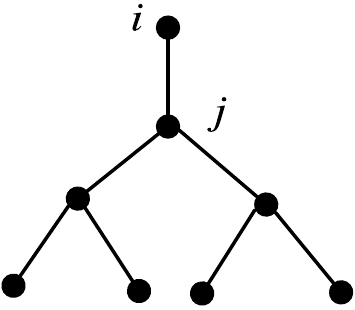}
\end{minipage}%
}%
\subfigure[\label{fig:cur-2}Positive Curvature]{
\begin{minipage}[t]{0.4\linewidth}
\centering
\includegraphics[width=0.4\textwidth]{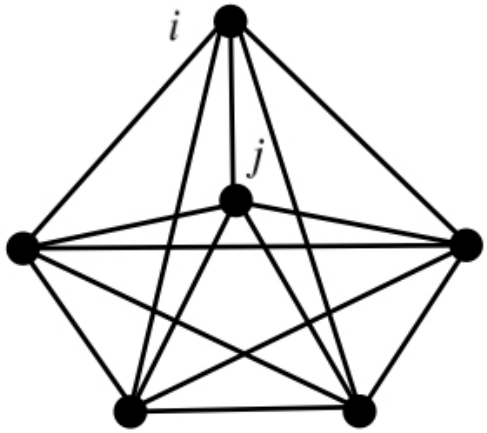}
\end{minipage}%
}%
\centering
\caption{\label{fig:curvature}Different regimes of curvatures on graphs. \textsl{Negative curvature will lead to over-squashing of information and graph bottleneck. Excessive positive curvature will increase information complexity.}}
\end{figure}

Figure \ref{fig:curvature} depicts two extreme cases that cause over-squashing of information. Let $i$ and $j$ be two adjacent nodes in the graph. There is only one path between node $i$ and node $j$ in Figure \ref{fig:cur-2}, and the number of 4-hop neighbors of node $i$ is $2^4$ which grows exponentially with the order. The information of the 4-hop neighbors of node $i$ can only be transmitted to node $j$ through node $i$, so node $j$ will cause the problem of information bottleneck in the process of information aggregation. While there are multiple paths between node $i$ and node $j$ in Figure \ref{fig:cur-1}. the 4-hop neighbors of node $i$ are all connected to node $j$, and the information of the 4-hop neighbors of node $i$ can be directly or indirectly transmitted to the node $j$. Therefore, there is no problem of information compression. However, it should be noted that the information aggregation of too many paths will increase the information complexity and introduce noise to a certain extent.

The Forman curvature \cite{forman2003bochner} is a tool used to measure over-squashing of information. Let $d_i$ and $d_j$ denote the degree of $i$ and $j$, $triangles_{<i,j>}$ denote triangles formed by $i$ and $j$ as two vertices. The Forman curvature is given by 
\begin{equation}
F(i,j)=4-d_i-d_j+3|triangles_{<i,j>}|.
\end{equation}

Figures \ref{fig:cur-1} and \ref{fig:cur-2} respectively show the case of negative and positive curvature. Shown as Figure \ref{fig:cur-2}, an open triangle is formed between every three nodes, that is, $|triangles_{<i,j>}|=0$. Let $triples_{<i,j>}$ be open triangles formed with $i$ and $j$ as two vertices, then 
\begin{equation}
d_i+d_j = \vert{triples_{<i,j>}}\vert + 2\vert{triangles_{<i,j>}}\vert + 2.
\end{equation}

So the formula of $F(i,j)$ can be replaced by
\begin{equation}
\label{F_ij}
F(i,j) = 2 - \vert{triples_{<i,j>}}\vert + \vert{triangles_{<i,j>}}\vert.
\end{equation}

This shows that $F(i,j)$ decreases as the number of open triangles increases, causing the problem of negative curvature. 

On the other hand, there are edges between all nodes in Figure \ref{fig:cur-1}, that is, a closed triangle is formed between every three nodes. According to the formula of $F(i,j)$, $d_i$ is offset with one of $triangles_{<i,j>}$, and so is $d_j$. Finally, $F(i,j)$ is still determined by $triangles_{<i,j>}$. This shows that $F(i,j)$ increases as the number of triangles increases, forming a distribution with positive curvature. 

From the above analysis, it can be seen that curvature is related to the number of triangles and open triangles in graph, which is similar to the graph clustering coefficient \cite{watts1998collective} to a certain extent. The difference between them is that the Forman curvature $F(i,j)$ measures the edge density between two nodes, while the clustering coefficient measures the edge density of a single node. Assuming that the number of closed triangles connected to vertex $i$ is $triangles_i$, and the number of open triangles connected to vertex $i$ is $triples_i$, the clustering coefficient is given by 
\begin{equation}
\label{C_i}C_i=\frac{|triangles_i|}{|triples_i|+|triangles_i|}.
\end{equation}
According to the definitions of $triangles_i$ and $triples_i$, the relationships between $\vert{triangles_i}\vert$ and $\vert{triangles_{<i,j>}}\vert$, and $\vert{triples_i}\vert$ and $\vert{triples_{<i,j>}}\vert$ satisfy
\begin{equation}
\vert{triangles_i}\vert = \sum_{l\in\mathcal{N}_i}\vert{triangles_{<i,j>}}\vert
\end{equation}
and 
\begin{equation}
\vert{triples_i}\vert = \sum_{j\in\mathcal{N}_i}\vert{triples_{<i,j>}}\vert,
\end{equation}
respectively, where $l$ is the neighboring node of the node $i$ and $\mathcal{N}_i$ is the set of neighboring nodes. Considering the Forman curvature $F(i,j)$ given in (\ref{F_ij}) and the clustering coefficient, (\ref{C_i}) can be rewritten as
\begin{equation}
\begin{aligned}
C_i = &\frac{\vert{triangles_i}\vert}{\vert{triangles_i}\vert+\vert{triples_i}\vert}\nonumber\\
= &\frac{\sum_{j\in\mathcal{N}_i}\vert{triangles_{<i,j>}}\vert}{\sum_{j\in\mathcal{N}_i}\vert{triangles_{<i,j>}}\vert+\sum_{j\in\mathcal{N}_i}\vert{triples_{<i,j>}}\vert}\nonumber\\
= &1 - \frac{\sum_{j\in\mathcal{N}_i}\vert{triples_{<i,j>}}\vert}{\sum_{j\in\mathcal{N}_i}F(i,j)+\sum_{j\in\mathcal{N}_i}(2\vert{triples_{<i,j>}}\vert-2)}.    
\end{aligned}
\end{equation}

To sum up, the clustering coefficient and the Forman curvature are positively correlated. They can both be used to measure the density of graph, and they are affected by the number of triangles and open triangles in graph. Therefore, triangle augmentation strategies are proposed to change the distribution of the Forman curvature and the clustering coefficient by adding and removing edges based on triangle.

Next, the effect of triangle augmentation on the clustering coefficient and the Forman curvature is demonstrated. (\ref{C_i}) can be simplified as 
\begin{equation}
C_i = \frac{1}{1+\frac{\vert{triples_i}\vert}{\vert{triangles_i}\vert}}.
\end{equation}

Let $x$ be the number of newly added open triangles and $y$ be the number of newly added triangles in triangle augmentation. When setting the same number of added and removed edges, if the specific graph structure is not considered, it should be $x=y$. But for sparse networks, $\vert{triples_i}\vert > \vert{triangles_i}\vert$. In this case, $x \leq y$. $C^{\prime}_i$ that denotes the clustering coefficient after triangle augmentation is given by
\begin{equation}
\label{C_i2}C^{\prime}_i=\frac{1}{1+\frac{\vert{triples_i}\vert+x}{\vert{triangles_i}\vert+y}}
\end{equation}
Considering 
\begin{equation}
\frac{\vert{triples_i}\vert+x}{\vert{triangles_i}\vert+y} < \frac{\vert{triples_i}\vert}{\vert{triangles_i}\vert}
\end{equation}
it results in
\begin{equation}
C^{\prime}_i > C_i.
\end{equation}

In summary, the feasibility and effectiveness of triangle augmentation are demonstrated rigorously. Triangle augmentation increases the clustering coefficient and the Forman curvature appropriately by adding edges with negative curvature and removing edges with excessive positive curvature, so as to alleviate the problem of information compression and improve performance.

\subsection{Loss Function}

The training data contains labeled data and unlabeled data. In order to effectively combine both, the loss function consists of two parts, namely $loss_L$ and $loss_U$. $loss_L$ is the cross entropy loss for supervising the training of labeled data, which is used to calculate the difference between the result distribution of labeled data and the distribution of ground-truth labels. $loss_U$ is the consistency regularization loss for supervising unlabeled data training, which is used to calculate the minimum mean square error of the result distribution of unlabeled data before and after DA, so as to ensure data consistency. The expression of the loss function is given as follows:
\begin{equation}
loss = loss_L + \lambda_U loss_U
\end{equation}

\begin{equation}
loss_L = -\frac{1}{\begin{vmatrix}Y_L\end{vmatrix}}\sum_{l\in{Y_L}}\sum_{n=1}^N I(y_l=c)logZ_{lc}
\end{equation}

\begin{equation}
loss_U = \frac{1}{\begin{vmatrix}Y_U\end{vmatrix}}\sum_{u\in{Y_U}}\begin{Vmatrix}\bar{Z}^{'}_u-Z_u\end{Vmatrix}
\end{equation}

$\lambda_U$ is used to adjust the contribution ratio of unlabeled data. $\begin{vmatrix}Y_L\end{vmatrix}$ and $\begin{vmatrix}Y_U\end{vmatrix}$ denote the number of labeled and unlabeled data, respectively. $C$ is the number of label classes. $y_l$ is the true label of the $l$-th node and $Z_lc$ denotes the vector of the $l$-th node with label class c. $Z_u$ denotes the vector of the $u$-th unlabeled node after DA and $\bar{Z}^{'}_u$ denotes the vector of the $u$-th unlabeled node after DA.

Various DA Strategies feature exchange, triangle based edge adding and removing are used for unlabeled data in training. If $loss_U$ simply uses the minimum mean squared error above, it requires different DA strategies to be applied on the same graph, which leads to mutual interference between different augmentation strategies as NodeAug mentioned. For example, a node is augmented by feature exchange, triangle based edge adding and removing at the same time, which makes the data fluctuate greatly. Not only the attributes of the node itself change, but also the distribution of surrounding nodes. Therefore, in order to execute node augmentation and triangle augmentation in parallel without affecting each other and integrate effectively, we further optimize $loss_U$.

Firstly, calculate the overall distribution of each class for unlabeled data by taking the mean of all vectors. 
\begin{equation}
\bar{Z}_u = \frac{1}{\begin{vmatrix}Y_U\end{vmatrix}}\sum_{u\in{Y_U}}Z_u
\end{equation}

Secondly, sharpening is applied to the result of multiple DAs to get the true distribution of results for each unlabeled node. The true distribution probability of the $u$-th unlabeled node with label class $j$ should be 
\begin{equation}
\bar{Z}^{'}_{uj} = \frac{\bar{Z}^{\frac{1}{T}}_{uj}}{\sum_{c=0}^{C-1}\bar{Z}^{\frac{1}{T}}_{uc}}(0\leq j \leq{C-1)}
\end{equation}

where $T\in (0,1]$ is a sharpness coefficient to control the sharpness of the distribution of label classes. As $T$ approaches 0, the sharpened label distribution will approach a one-hot distribution.

Finally, obtain the resulting matrix $\bar{Z}^{'}_{u}$ of unlabeled nodes, in which the results of different DA strategies have more obvious differences in the probability of each classification. And then $\bar{Z}^{'}_{u}$ is input to $loss_U$.

\section{\label{experiments}Experiments}
In this section, we first present the experimental settings including datasets, comparison methods, evaluation metrics and parameter settings.
Then, we conduct comprehensive experiments on three widely used benchmark datasets and one newly collected industrial dataset to answer the following questions:
\begin{itemize}
\item{
Q1: How does HG-MDA perform on heterogeneous graph of SSGL comparing with the state-of-the-art methods?
}
\item{
Q2: How does the three key modules, i.e., the feature exchange of node augmentation module, the edge adding based triangle augmentation module, and the edge removing based triangle augmentation module, contribute to HG-MDA's performance?}
\item{
Q3: Comparing the HG-MDA method and the original method, how does it affect the node embedding training?}
\item{
Q4: What specific benefit does the proposed triangle augmentation bring to the graph structure (distribution of nodes)?}
\item{
Q5: How do the setting of key hyper-parameters make differences to HG-MDA's performance?}
\end{itemize}

\subsubsection{Dataset Descriptions}
To make the evaluation results general, three benchmark datasets: DBLP \cite{DBLP:journals/corr/abs-2005-11079}, ACM \cite{2019HAN} and a sub-graph of ogbn-mag (sub-ogbn) \cite{2020Microsoft} and an industrial dataset MB are used to evaluate the performance. Table ~\ref{tab:data} summarizes the statistics of all datasets. They are heterogeneous information networks. Specifically, the former three are citation networks, each of which refers to subject areas of ACM conference or Computer Science arXiv related papers and authors, such as database, wireless communication, data mining, etc. Note that sub-ogbn is a sub-graph of three categories sampled from ogbn-mag. The MB dataset is more challenging dataset collected from Ant Group, describing the user interactions in financial products such as friends, communication, and transfer transactions. For each dataset, we randomly split according to 24\% for training, 6\% for validation and 70\% for test. And early stopping with patience of 30 is set, i.e., it stop training if the loss does not decrease for 30 consecutive epochs. 

\subsubsection{Comparison Methods}
We compare HG-MDA with the following methods of GCA, GRACE, BGRL, G-BT, GTN, RGCN, HAN, HGT, GRAND and NASA for comparison on the node classification task of ACM dataset. The former four methods are GCL for semi-supervised learning, the next four methods are Semi-supervised HGNNs and the latter two comparison methods are SSGL-DA.

\begin{itemize}
\item \textbf{GCL methods}
\begin{itemize}
\item \textbf{GCA} \cite{2021GCA} - a self-supervised pre-training method of graph representation learning for node-level contrastive objectives, which uses adaptive data augmentation such as edge removing and feature masking for unimportant edges and features, and uses InfoNCE loss for contrastive learning.
\item \textbf{GRACE} \cite{2020Deep} - a self-supervised pre-training method of graph representation learning, which uses data augmentation through random edge removing and feature masking, and InfoNCE loss for contrastive learning.
\item \textbf{BGRL} \cite{2021Bootstrapped} - a self-supervised pre-training method of graph representation learning that does not require negative pairs. Data augmentation is performed through random edge removing and feature masking, and the Bootstrapping Latent loss is used for contrastive learning.
\item \textbf{G-BT} \cite{2021GBT} - a self-supervised pre-training method of graph representation learning, which uses data augmentation through random edge removing and feature masking, and Barlow Twins loss for contrastive learning.
\end{itemize}
\item \textbf{Semi-supervised HGNN methods \cite{wang2022survey}}
\begin{itemize}
\item \textbf{GTN} \cite{2019GTN} - a HGNN that can automatically discover valuable meta-paths, which uses a soft sub-graph selection and matrix multiplication step to generate meta-path neighbor graphs, and then encodes the graphs by GCNs.
\item \textbf{RGCN} \cite{2018RGCN} - a HGNN dealing with multi-edge types, in which the convolution can be interpreted as a weighted sum of ordinary graph convolution with different edge types.
\item \textbf{HAN} \cite{2019HAN} - a GAT-based HGNN, which uses a hierarchical attention mechanism to capture both node-level and semantic-level importance.
\item \textbf{HGT} \cite{2020HGT} – a transformer-based HGNN that learns specialized heterogeneous attention for different types of nodes and edges.
\end{itemize}
\item \textbf{SSGL-DA methods \cite{wang2022survey}}
\begin{itemize}
\item \textbf{GRAND} \cite{feng2020graph} - A semi-supervised learning framework via random propagation of features and consistent regularization strategies, where in each node's features can be randomly dropped either partially (dropout) or entirely.
\item \textbf{NASA} \cite{2022NASA} - A semi-supervised learning framework consisting of graph augmentation and regularization, where adjacent nodes are treated as special augmentations and interference are introduced by replacing adjacent nodes with distant neighbors.
\end{itemize}
\end{itemize}

\subsubsection{Evaluation Metrics}
We employ two widely used metrics, i.e., $Macro-F1, Micro-F1$, to evaluate the performance of all methods. In order to ensure the stability of the evaluation, the final evaluation score is the average of the results with 5 times running. For all the two metrics, higher values indicate better performance. 

Let $M$ denotes the number of categories. $P_i$ denotes precision index of the $i-th$ class and $R_i$ denotes precision index of the $i-th$ class. The calculation formulas of $Macro-F1$ are:

\[Macro\-F1 = \frac{\sum_{i=1}^M F1\-score_i}{M}\]
\[F1\-score_i = \frac{2P_i R_i}{P_i+R_i}\]

$TP_i$ denotes True Positive of the $i-th$ class that the positive class is judged as the positive class. $FN_i$ denotes False Negative of the $i-th$ class that the positive class is judged as the negative class. And $FP_i$ denotes False Positive of the $i-th$ class that negative class is judged as positive class. The calculation formulas of $Micro-F1$ are:

\[Micro-F1 = \frac{2P_M R_M}{P_M+R_M}\]
\[P_M = \frac{\sum_{i=1}^M TP_i}{\sum_{i=1}^M TP_i+\sum_{i=1}^M FN_i}\]
\[R_M = \frac{\sum_{i=1}^M TP_i}{\sum_{i=1}^M TP_i+\sum_{i=1}^M FP_i}\]

\subsubsection{Parameter Settings}

For the hyper-parameters of HG-MDA model, we set the number of aggregation layers as 5 and the number of hidden units as 64. Model are optimized via the AdamW optimizer. For meta-relation-based attention, the number of attention heads is set as 8. In the loss function, the harmonic coefficient $\lambda_U$ is set as 0.5 to adjust the contribution ratio of labeled loss and unlabeled loss. The hyper-parameters of other models are consistent with their original papers. For each model, we train it for 300 epochs and select the model with the lowest validation loss as the reporting model. And prediction results of each experiment are averaged after the experiment runs 5 times respectively.

\begin{table}
\centering
\caption{\label{tab:data}The Statistics of All Datasets.}
\resizebox{\textwidth}{!}{
\begin{tabular}{@{}c|cccccccc@{}}
\toprule
\textbf{Dataset} &
  \textbf{Domain} &
  \textbf{Nodes} &
  \textbf{Nodes Types} &
  \textbf{Edges} &
  \textbf{Edges Types} &
  \textbf{Feature dimension} &
  \textbf{target} &
  \textbf{Classes} \\ \midrule
ACM      & Citation networks & 9,040 & 3 & 31,291 & 5 & 1,903 & paper & 3  \\
DBLP     & Citation networks & 26,128 & 4 & 239,566 & 6 & 334  & author  & 4 \\
sub-ogbn & Citation networks & 21,949 & 4 & 112,646 & 4 & 128  & paper  & 4 \\
MB       & Social networks   & 40,149 & 2 & 71,305 & 11 & 68  & user   & 2  \\ \bottomrule
\end{tabular}
}
\end{table}

\subsection{Performance of Methods (for Q1)}
In this subsection, we conduct experiments in two different dimensions to answer Q1. The first part is comparison experiments between HG-MDA and other benchmark models. The second part is expansion experiments that the proposed Multi-level Data Augmentation (MDA) strategies are applied to multi basic frameworks of HGNNs to verify the general applicability of the proposed MDA strategies. 

\subsubsection{Comparison Experiments}
Our proposed framework is evaluated in the semi-supervised learning
setting on node classification on the four datasets. All methods follow exactly the same experimental procedure, such as feature processing, data splits and ratio of unlabeled data. Table~\ref{tab:results} reports the Macro-F1 and Micro-F1 results of all methods. HG-MDA takes HGT as the basic framework and applies the multi-level heterogeneous DA strategies. Since HG-MDA consists of HGNN and DA, where the augmentation strategies refer to GCL methods, HG-MDA is compared with the current popular GCL, HGNNs and SSGL-DA methods respectively. From then we have several observations:

\textbf{1) Comparison with GCL methods for semi-supervised learning:} (1) HG-MDA outperforms all compared GCL methods by an average of 24.1\% on the Macro-F1 metric and 11.1\% on the Micro-F1 metric on the four datasets. DA strategies in these GCL methods are generally applicable in homogeneous graph, while DA strategies in HG-MDA are designed for the characteristics of heterogeneous graph, indicating that the heterogeneous information in DA can improve the performance of the model. (2) on the class-imbalanced datasets, i.e., DBLP and sub-ogbn, GCL methods have large biases in the prediction results, which are reflected in lower Macro-F1 scores and higher Micro-F1 scores. But there is no bias problem in HG-MDA. The reason might be that GCL is node representation learning for the whole graph that does not distinguish node types, so that the model does not pay attention to key information. But HG-DMA focuses on the target node type, distinguishes the nodes attribute information and the different preferences of other nodes and edge types for data enhancement, so that the target node representation has richer interaction semantics.

\textbf{2) Comparison with Semi-supervised HGNNs methods:} (1) HG-MDA outperforms all compared HGNNs methods by an average of 18.5\% on the Macro-F1 metric and 11.6\% on the Micro-F1 metric on the four datasets. Compared with the HGNNs that only supervises the labeled data, HG-MDA can further improve the model performance by additionally using unlabeled data for training. (2) HG-MDA outperforms the basic method HGT by an average of 4.6\% on the Macro-F1 metric and 3.2\% on the Micro-F1 metric on the four datasets, which indicates that DA and consistent learning is beneficial to learn target nodes representations.

\textbf{3) Comparison with SSGL-DA methods:} (1) HG-MDA has an average improvement of 8.7\% on the Macro-F1 metric and 5.1\% on the Micro-F1 metric on the four datasets. The improvement may be brought about by two aspects. First, unlike the HG-MDA proposed specifically for heterogeneous graphs, the SSGL-DA method is mainly applicable to homogeneous graphs, which indicates that capturing the heterogeneity of information on node attributes and structure is helpful in HG-MDA. Second, the DA strategies in SSGL-DA methods are random, while HG-MDA selects nodes and edges according to relevant indicators, i.e., meta-relation-based attention, feature similarity and $AA$ index, which can bring more effective information to model training. (2) compared with HG-MDA and other models, The performance of GRAND and NASA on real-world industrial dataset MB is not ideal. This is because the dimensions of node features are less and the connections between nodes are sparse on the MB dataset. This observation validates that HG-MDA can handle sparse scenes more efficiently and achieve satisfactory results than all other methods.

\begin{table}
\centering
\setlength{\tabcolsep}{2pt}
\caption{\label{tab:results}Semi-supervised Node Classification Results.}
\resizebox{\textwidth}{!}{
\begin{threeparttable}

\begin{tabular}{@{}l|cc|cc|cc|cc@{}}
\toprule
\multicolumn{1}{c|}{} & \multicolumn{2}{c|}{ACM} & \multicolumn{2}{c|}{DBLP} & \multicolumn{2}{c|}{Sub-ogbn} & \multicolumn{2}{c}{MB} \\ \midrule
\multicolumn{1}{c|}{Metric} & Macro-F1 & Micro-F1 & Macro-F1 & Micro-F1 & Macro-F1 & Micro-F1 & Macro-F1 & Micro-F1 \\ \midrule
GCA & 0.687$\pm$0.024 & 0.690$\pm$0.021 & 0.568$\pm$0.018 & 0.881$\pm$0.019 & 0.525$\pm$0.012 & 0.810$\pm$0.013 & 0.729$\pm$0.012 & 0.773$\pm$0.012 \\
GRACE & 0.688$\pm$0.020 & 0.690$\pm$0.019 & 0.548$\pm$0.014 & 0.874$\pm$0.014 & 0.432$\pm$0.018 & 0.804$\pm$0.020 & \textbf{0.745$\pm$0.010} & \textbf{0.783$\pm$0.011} \\
BGRL & 0.737$\pm$0.014 & 0.740$\pm$0.014 & 0.676$\pm$0.011 & 0.905$\pm$0.012 & 0.665$\pm$0.018 & 0.865$\pm$0.016 & 0.732$\pm$0.012 & 0.769$\pm$0.012 \\
G-BT & 0.684$\pm$0.010 & 0.687$\pm$0.010 & 0.614$\pm$0.015 & 0.889$\pm$0.016 & 0.596$\pm$0.011 & 0.841$\pm$0.010 & 0.687$\pm$0.013 & 0.740$\pm$0.013 \\ \midrule
GTN & 0.736$\pm$0.018 & 0.739$\pm$0.017 & 0.765$\pm$0.022 & 0.795$\pm$0.020 & 0.712$\pm$0.033 & 0.765$\pm$0.009 & 0.644$\pm$0.023 & 0.682$\pm$0.028 \\
RGCN & 0.830$\pm$0.014 & 0.827$\pm$0.014 & 0.914$\pm$0.010 & 0.919$\pm$0.010 & 0.646$\pm$0.043 & 0.927$\pm$0.009 & 0.676$\pm$0.010 & 0.704$\pm$0.025 \\
HAN & 0.887$\pm$0.014 & 0.886$\pm$0.014 & 0.900$\pm$0.014 & 0.904$\pm$0.014 & 0.476$\pm$0.004 & 0.908$\pm$0.013 & 0.680$\pm$0.014 & 0.696$\pm$0.020 \\
HGT & 0.908$\pm$0.024 & 0.907$\pm$0.018 & \textbf{0.928$\pm$0.013} & \textbf{0.933$\pm$0.013} & \textbf{0.815$\pm$0.012} & 0.923$\pm$0.011 & 0.707$\pm$0.022 & 0.740$\pm$0.021 \\ \midrule
GRAND & \textbf{0.931$\pm$0.007} & \textbf{0.930$\pm$0.007} & 0.902$\pm$0.014 & 0.905$\pm$0.012 & 0.740$\pm$0.005 & \textbf{0.937$\pm$0.013} & 0.612$\pm$0.014 & 0.665$\pm$0.020 \\
NASA & 0.928$\pm$0.008 & 0.927$\pm$0.008 & 0.904$\pm$0.013 & 0.911$\pm$0.015 & 0.718$\pm$0.008 & 0.907$\pm$0.012 & 0.654$\pm$0.015 & 0.680$\pm$0.018 \\ \midrule
HG-MDA & \textbf{0.934$\pm$0.012} & \textbf{0.934$\pm$0.011} & \textbf{0.948$\pm$0.016} & \textbf{0.951$\pm$0.014} & \textbf{0.878$\pm$0.011} & \textbf{0.942$\pm$0.010} & \textbf{0.781$\pm$0.009} & \textbf{0.802$\pm$0.015} \\ \bottomrule
\end{tabular}
\begin{tablenotes}
\item Vacant positions ("-") indicates not applicable.
\end{tablenotes}

\end{threeparttable}
}
\end{table}

\subsubsection{Expansion Experiments}
We take the GNN methods, i.e., GTN, RGCN, HAN, and HGT, as the baselines, the augmentation strategies proposed in this paper are applied to these methods to obtain the corresponding models GTN-MDA, RGCN-MDA, HAN-MDA, and HG-MDA respectively. Comparing the extended models with baselines, we can also draw two conclusions from Table~\ref{tab:results}.

1) the proposed augmentation strategies can be universally applied to various HGNNs and bring gains. They enhances the strong GTN, RGCN, HAN, and HGT on average by 4.26\%, 1.3\%, 0.9\%, 4.6\% on the Macro-F1 metric and 3.8\%, 1.3\%, 1\%, 3.2\% on the Micro-F1 metric on all datasets.

2) Among MDA-based methods, due to the excellent performance of HGT and the perfect combination of HGT and augmentation strategies, HG-MDA achieved the best performance compared with GTN-MDA, RGCN-MDA, HAN-MDA.

\begin{table}
\centering
\caption{\label{tab:results2}Expansion Experiments of HG-MDA.}
\resizebox{\textwidth}{!}{
\begin{threeparttable}

\begin{tabular}{@{}l|cc|cc|cc|cc@{}}
\toprule
\multicolumn{1}{c|}{} & \multicolumn{2}{c|}{ACM} & \multicolumn{2}{c|}{DBLP} & \multicolumn{2}{c|}{Sub-ogbn} & \multicolumn{2}{c}{MB} \\ \midrule
\multicolumn{1}{c|}{Metric} & Macro-F1 & Micro-F1 & Macro-F1 & Micro-F1 & Macro-F1 & Micro-F1 & Macro-F1 & Micro-F1 \\ \midrule
GTN & 0.736$\pm$0.018 & 0.739$\pm$0.017 & 0.765$\pm$0.022 & 0.795$\pm$0.020 & 0.712$\pm$0.033 & 0.765$\pm$0.009 & 0.644$\pm$0.023 & 0.682$\pm$0.028 \\
RGCN & 0.830$\pm$0.014 & 0.827$\pm$0.014 & 0.914$\pm$0.010 & 0.919$\pm$0.010 & 0.646$\pm$0.043 & \textbf{0.927$\pm$0.009} & 0.676$\pm$0.010 & 0.704$\pm$0.025 \\
HAN & 0.887$\pm$0.014 & 0.886$\pm$0.014 & 0.900$\pm$0.014 & 0.904$\pm$0.014 & 0.476$\pm$0.004 & 0.908$\pm$0.013 & 0.680$\pm$0.014 & 0.696$\pm$0.020 \\
HGT & \textbf{0.908$\pm$0.024} & \textbf{0.907$\pm$0.018} & \textbf{0.928$\pm$0.013} & \textbf{0.933$\pm$0.013} & \textbf{0.815$\pm$0.012} & 0.923$\pm$0.011 & \textbf{0.707$\pm$0.022} & \textbf{0.740$\pm$0.021} \\ \midrule
GTN-MDA & 0.568$\pm$0.020 & 0.569$\pm$0.020 & 0.387$\pm$0.022 & 0.426$\pm$0.022 & 0.756$\pm$0.020 & 0.809$\pm$0.007 & 0.750$\pm$0.019 & 0.766$\pm$0.016 \\
RGCN-MDA & 0.841$\pm$0.014 & 0.839$\pm$0.014 & 0.930$\pm$0.006 & 0.935$\pm$0.006 & 0.662$\pm$0.035 & 0.930$\pm$0.009 & 0.687$\pm$0.010 & 0.724$\pm$0.025 \\
HAN-MDA & 0.889$\pm$0.011 & 0.888$\pm$0.011 & 0.912$\pm$0.013 & 0.916$\pm$0.012 & 0.477$\pm$0.001 & 0.912$\pm$0.003 & 0.720$\pm$0.012 & 0.726$\pm$0.023 \\
HG-MDA & \textbf{0.934$\pm$0.012} & \textbf{0.934$\pm$0.011} & \textbf{0.948$\pm$0.016} & \textbf{0.951$\pm$0.014} & \textbf{0.878$\pm$0.011} & \textbf{0.942$\pm$0.010} & \textbf{0.781$\pm$0.009} & \textbf{0.802$\pm$0.015} \\ \bottomrule
\end{tabular}
\begin{tablenotes}
\item  The augmentation strategies of HG-MDA is extended to other heterogeneous graph neural network frameworks to verify the effectiveness and portability of the strategies proposed in this paper.
\end{tablenotes}
\end{threeparttable}
}

\end{table}

\subsection{Ablation Studies (for Q2)}

To answer Q2, the ablation studies with four simplified versions of HG-MDA are conducted on two essential tasks: node classification and node clustering to examine the contribution of three DA strategies and the performance of the final integrated model HG-MDA: (1) \textbf{Baseline}: HG-MDA without any augmentation strategies and with cross entropy loss, i.e., HGT is used as the basic framework, (2) \textbf{+ Feature Exchange}: baseline add the node augmentation strategy of feature exchange, (3) \textbf{+ Triangle based Edge Adding}: baseline add the augmentation strategy of triangle based edge adding, (4) \textbf{+ Triangle based Edge Removing}: baseline add the augmentation strategy of triangle based edge removing, and (5) \textbf{HG-MDA with consistency loss}: the final HG-MDA with the three augmentation strategies comprehensively and the optimized consistency loss function. The sharpness coefficient $T$ of the optimized loss function is set as 0.2 and the augmentation ratio $K$ is set to 0.5, which will be shown as the best hyper-parameters in the Section \ref{key-parameters}.

Table~\ref{tab:ablation} summarizes the ablation studies for two tasks of node classification and node clustering. It should be noted that there are only 1-hop nodes but no 2-hop nodes for the target nodes on the DBLP dataset, that is the number of triangles is 0, which is shown in Table~\ref{tab:original_triangles}. Therefore, the experiments of edge adding and edge removing based on triangle can not be performed on the DBLP dataset. The corresponding experiment of HG-MDA with consistency loss on the DBLP dataset is performed on the node augmentation of different augmentation ratio $K$. 

The following conclusions can be drawn from Table~\ref{tab:ablation}: 

1) the three DA strategies all bring performance improvements to the model. \textbf{(+ Feature Exchange)} bring improvement than the baseline, indicating that the diversity provided by node augmentation is helpful to model training. Both \textbf{(+ Triangle based Edge Adding)} and \textbf{(+ Triangle based Edge Removing)} also show that triangle augmentation can provide information gain for the model, even larger than node augmentation. From the analysis 
in Section \label{geometric}, it can be seen that triangle augmentation changes the distribution of triangles and open triangles in the graph, thus alleviating the information compression problem caused by negative curvature and learning better node representations.

2) Comparing the results of \textbf{(+ Triangle based Edge Adding)} and \textbf{+ Triangle based Edge Removing)}, \textbf{(+ Triangle based Edge Adding)} has a better performance on overall clustering tasks, while \textbf{(+ Triangle based Edge Removing)} performs better on overall classification tasks. Triangle based edge removing removes some edges with weak information transmission to 
reduces noise in the graph and simplifies relationships between nodes, which enhances the differences between nodes to some extent. Therefore, it has better performance in node classification. Triangle based edge adding adds edges with strong information between highly correlated nodes, making the graph structure more compact and the nodes closer together. So it has better performance in node clustering.

3) Furthermore, from Table ~\ref{tab:results} and Table ~\ref{tab:ablation}, not only HG-MDA, but also the model with triangle based edge adding or with triangle based edge removing outperform most of the benchmark methods, indicating that triangle augmentation is the good way to augment graphs.

4) \textbf{(HG-MDA with consistency loss)} that integrated with multiple DAs and the consistency loss achieves the best performance on both tasks of all four datasets, which validates that our proposed method is effective. Multi-level data enhancement can learn node attributes and structural information separately, and finally combined to play an excellent performance.

\begin{table}
\centering
\caption{\label{tab:ablation} Ablation Studies for HG-MDA.}
\resizebox{\textwidth}{!}{
\begin{threeparttable}
\begin{tabular}{@{}c|l|cc|cc|cc|cc@{}}
\toprule
Task & \multicolumn{1}{c|}{Component} & \multicolumn{2}{c|}{ACM} & \multicolumn{2}{c|}{DBLP} & \multicolumn{2}{c|}{Sub-ogbn} & \multicolumn{2}{c}{MB} \\ \midrule
\multirow{7}{*}{Node Classification} & \multicolumn{1}{c|}{Metric} & Macro-F1 & Micro-F1 & Macro-F1 & Micro-F1 & Macro-F1 & Micro-F1 & Macro-F1 & Micro-F1 \\ \cmidrule(l){2-10} 
 & Baseline & 0.908$\pm$0.018 & 0.907$\pm$0.019 & 0.928$\pm$0.017 & 0.933$\pm$0.018 & 0.815$\pm$0.012 & 0.923$\pm$0.014 & 0.707$\pm$0.012 & 0.740$\pm$0.012 \\
 & + Feature Exchange & 0.915$\pm$0.013 & 0.914$\pm$0.014 & 0.933$\pm$0.015 & 0.937$\pm$0.015 & 0.844$\pm$0.012 & 0.931$\pm$0.012 & 0.712$\pm$0.011 & 0.712$\pm$0.010 \\
 & + Triangle based Edge Adding & 0.921$\pm$0.012 & 0.920$\pm$0.011 & - & - & 0.840$\pm$0.012 & 0.930$\pm$0.012 & 0.712$\pm$0.013 & 0.746$\pm$0.013 \\
 & + Triangle based Edge Removing & 0.929$\pm$0.012 & 0.928$\pm$0.012 & - & - & 0.842$\pm$0.010 & 0.934$\pm$0.009 & 0.698$\pm$0.012 & 0.733$\pm$0.012 \\
 & HG-MDA with consistency loss & \textbf{0.934$\pm$0.012} & \textbf{0.934$\pm$0.011} & \textbf{0.948$\pm$0.016} & \textbf{0.951$\pm$0.014} & \textbf{0.878$\pm$0.011} & \textbf{0.942$\pm$0.010} & \textbf{0.781$\pm$0.009} & \textbf{0.802$\pm$0.015} \\ \midrule
\multirow{7}{*}{Node Clustering} & \multicolumn{1}{c|}{Metric} & NMI & ARI & NMI & ARI & NMI & ARI & NMI & ARI \\ \cmidrule(l){2-10} 
 & Baseline & 0.697$\pm$8e-5 & 0.755$\pm$7e-5 & 0.733$\pm$7e-5 & 0.795$\pm$9e-5 & 0.575$\pm$5e-5 & 0.451$\pm$5e-5 & 0.099$\pm$2e-6 & 0.143$\pm$2e-6 \\
 & + Feature Exchange & 0.720$\pm$8e-5 & 0.773$\pm$8e-5 & 0.763$\pm$7e-5 & 0.821$\pm$6e-5 & 0.678$\pm$3e-5 & 0.704$\pm$3e-5 & 0.069$\pm$1e-6 & 0.107$\pm$1e-6 \\
 & + Triangle based Edge Adding & 0.759$\pm$7e-5 & 0.810$\pm$8e-5 & - & - & \textbf{0.722$\pm$4e-5} & \textbf{0.796$\pm$5e-5} & 0.150$\pm$2e-6 & 0.204$\pm$2e-6 \\
 & + Triangle based Edge Removing & 0.730$\pm$6e-5 & 0.776$\pm$7e-5 & - & - & 0.662$\pm$3e-5 & 0.645$\pm$3e-5 & 0.112$\pm$2e-6 & 0.172$\pm$1e-6 \\
 & HG-MDA with consistency loss & \textbf{0.760$\pm$6e-5} & \textbf{0.811$\pm$6e-5} & \textbf{0.793$\pm$4e-5} & \textbf{0.845$\pm$4e-5} & 0.678$\pm$2e-5 & 0.752$\pm$2e-5 & \textbf{0.194$\pm$1e-6} & \textbf{0.230$\pm$1e-6} \\ \bottomrule
\end{tabular}
\begin{tablenotes}
\item "+ feature exchange" means that the node augmentation - feature exchange is applied to baseline. "HG-MDA with base loss" meas that $loss_U$ uses MSE loss for consistency regularization in HG-MDA. In method 2,3,4, $loss_U$ also uses MSE loss. "HG-MDA with optimized loss" meas that $loss_U$ uses the optimizd loss proposed in our approach. Vacant positions ("-") indicates not applicable, because in the DBLP dataset, there are only first-order nodes and no second-order nodes for the target node, so it is impossible to add and delete edges based on triangles.
\end{tablenotes}
\end{threeparttable}
}

\end{table}

\subsection{Visualization of Node Clustering (for Q3)}

\begin{figure}[htbp]
\centering
\subfigure[\label{fig:Baseline} Baseline]{
\begin{minipage}[t]{0.5\linewidth}
\centering
\includegraphics[width=0.7\textwidth]{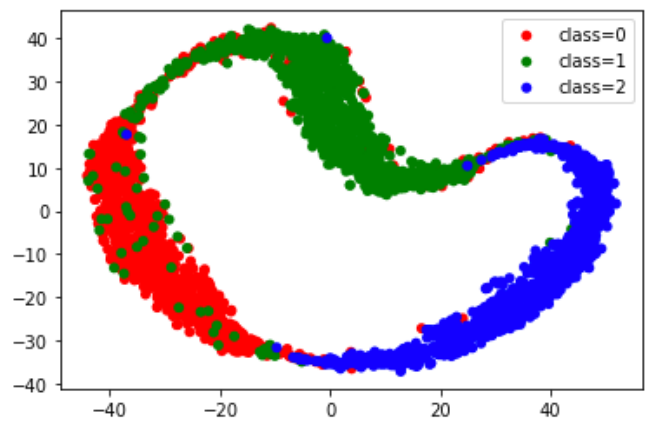}
\end{minipage}%
}%
\subfigure[\label{fig:HGMDA} HG-MDA]{
\begin{minipage}[t]{0.5\linewidth}
\centering
\includegraphics[width=0.7\textwidth]{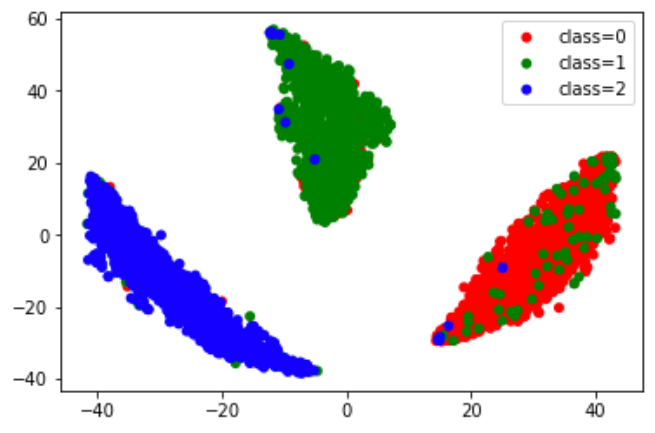}
\end{minipage}%
}%
\centering
\caption{\label{fig:visualization}The Visualization of node embedding of baseline and HG-MDA in the ACM dataset. \textsl{Colors denote the ground-truth class labels.}}
\end{figure}

To answer Q3, Figure \ref{fig:visualization} shows the visualization of the node embedding of the target nodes obtained by the baseline model and the HG-MDA model on the ACM dataset and compressed dimension by t-SNE to project on the 2D coordinates. There are three categories of target nodes on the ACM dataset. It can be seen from Figure \ref{fig:Baseline} that the node distribution of each category output by the baseline model still has a certain intersection, and the categories of some nodes are not easy to distinguish. However, in Figure \ref{fig:HGMDA}, there is a clear dividing line between the distribution of various categories of nodes in the HG-MDA model. It shows that the node embeddings inter each category are farther apart, while the node embeddings intra each category are more compact after MDA strategies.

\subsection{Benefits of Triangle Augmentation (for Q4)}

To answer Q4, the upper and lower parts of Table~\ref{tab:original_triangles} respectively show the changes in the distribution of triangles and open triangles before and after triangle augmentation. In terms of computational complexity and observation convenience, the clustering coefficient is simpler and easier to observe than the curvature. Therefore, in this subsection, the clustering coefficient is selected as the observation index to illustrate the change in the density of the graph structure before and after the triangle augmentation strategies.

In Table~\ref{tab:original_triangles}, compared with the distribution before triangle augmentation, triangle augmentation brings two benefits. First, the number of triangles increases, while the number of open triangles decreases. It means that triangle based edges adding and removing make the connections between nodes denser, thus conveying richer information. Second, from the $\Delta_{Mean}$, it can be seen that the mean of clustering coefficients in graph after the triangle augmentation respectively increase by 0.018, 0.009, and 0.068 on the ACM, sub-ogbn, and MB datasets. The increase in the mean of clustering coefficient indicates a corresponding increase in the Forman curvature of the entire graph, that is, the problem of information compression is alleviated by triangle augmentation to a certain degree. It should be noted that there is no triangle on the DBLP dataset, because there are only 1-hop nodes but no 2-hop nodes for the target nodes. 

\begin{figure}[htbp]
\centering
\subfigure[\label{fig:node_degrees}]{
\begin{minipage}[t]{0.5\linewidth}
\centering
\includegraphics[width=0.7\textwidth]{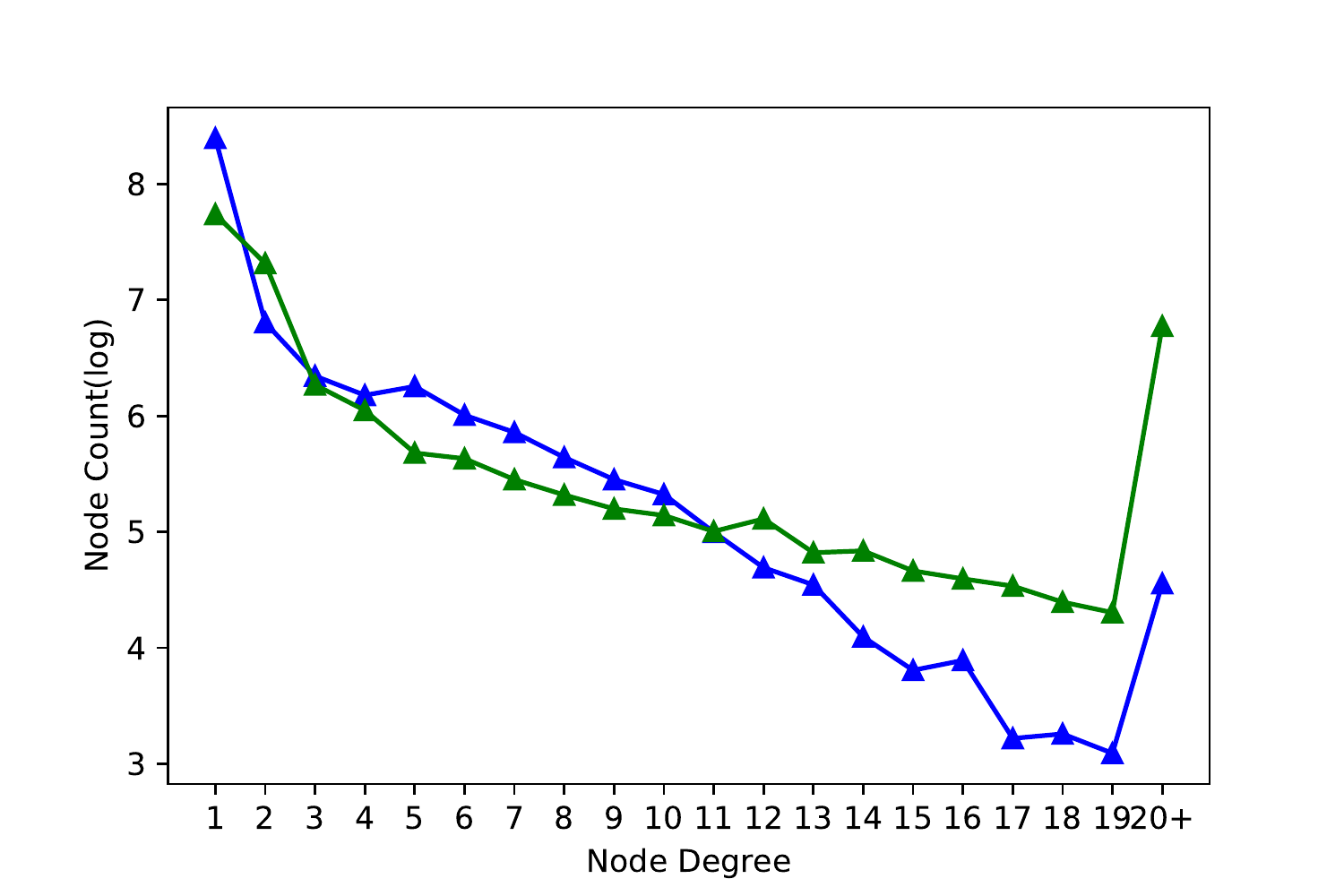}
\end{minipage}%
}%
\subfigure[\label{fig:ci}]{
\begin{minipage}[t]{0.5\linewidth}
\centering
\includegraphics[width=0.7\textwidth]{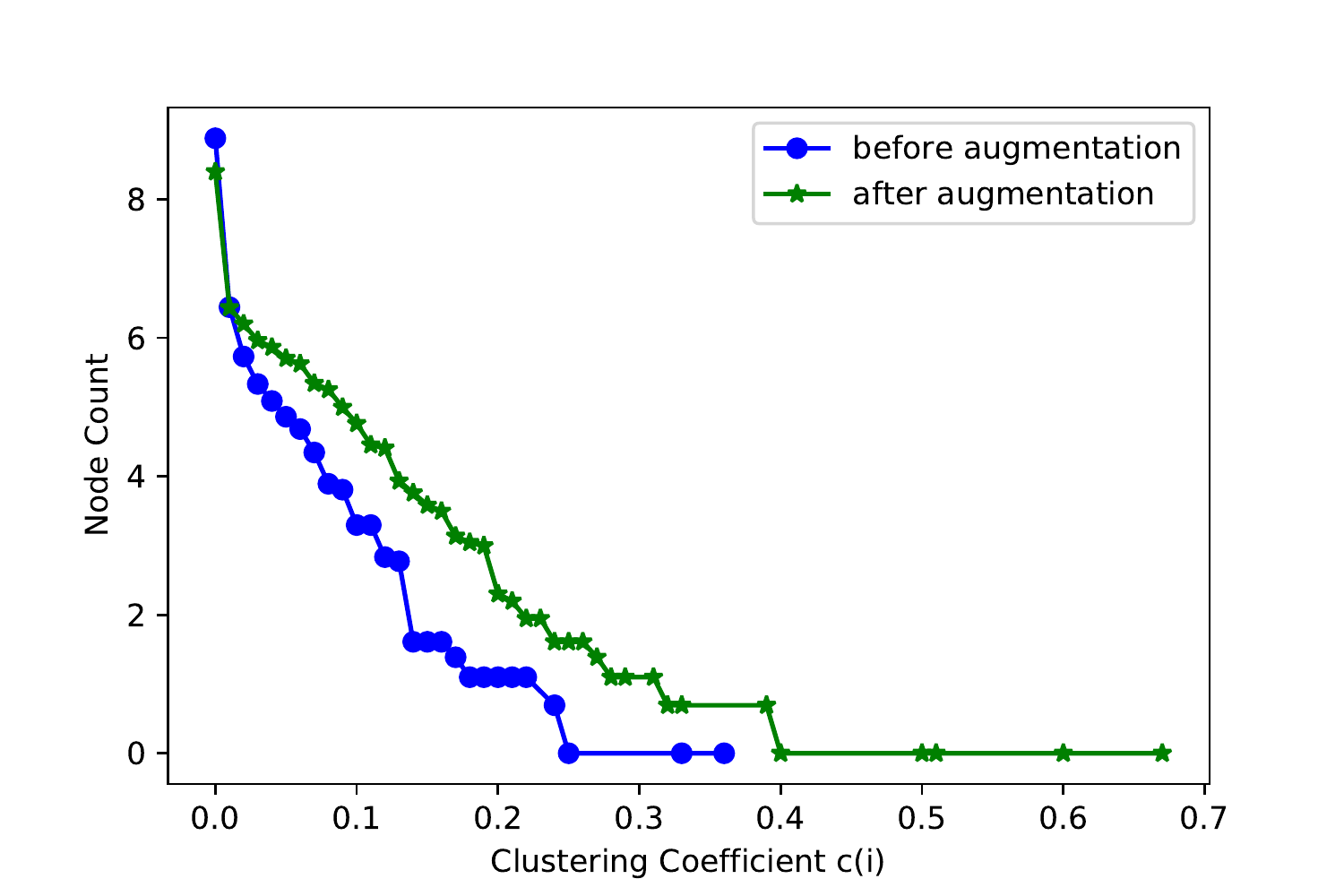}
\end{minipage}%
}
\caption{(a) The Change of Node Degrees Before and After Triangle Augmentations. \textsl{The x-axis represents the degree of nodes, and the y-axis represents the $log_{10}$ of nodes with corresponding degrees.} (b) The Change of $C_i$ Before and After Triangle Augmentations. \textsl{The x-axis represents the value of $C_i$ for node $i$, and the y-axis represents the $log_{10}$ of nodes with corresponding degrees.}}

\end{figure}

Furthermore, Figure \ref{fig:node_degrees} shows the distribution of node degrees before and after triangle augmentations. It can be seen that triangle augmentation can decrease the number of open triangles and increase the number of triangles which makes the distribution of node degrees flatter. and Figure \ref{fig:ci} shows the distribution of clustering coefficient for each node before and after triangle augmentations. The overall distribution of $C_i$ moves to the right, which shows that after triangle augmentation, the clustering coefficient of each node has increased. It is consistent with the conclusion obtained in Table ~\ref{tab:original_triangles}.

Clearly, This subsection verifies the geometric analysis in Section \ref{geometric}, triangle augmentation changes the distribution of triangles and open triangles in graph by adding and removing edges based on triangles to make the clustering coefficient and curvature increase, thereby alleviating the problem of information compression caused by negative curvature, optimizing the transmission of information in the network and bring performance gains.

\begin{table}
\centering
\caption{\label{tab:original_triangles}The Changes in The Distribution of Triangles and Open Triangles Before and After Triangle Augmentation Strategies.}
\begin{threeparttable}

\begin{tabular}{@{}l|cccc@{}}
\toprule
Dataset & \multicolumn{1}{l}{the number of open triangles} & \multicolumn{1}{l}{the number of triangles} & \multicolumn{1}{l}{$Mean(C_i)$} & \multicolumn{1}{l}{$\Delta_{Mean}$} \\ \midrule
ACM & 1254834 & 7055 & 0.008 & - \\
DBLP & 19644946 & 0 & 0 & - \\
sub-ogbn & 8912884 & 13766 & 0.003 & - \\
MB & 231043 & 13114 & 0.117 & - \\ \bottomrule
ACM & 1518373 & 37291 & 0.026 & 0.018 \\
DBLP & 19644946 & 0 & 0 & - \\
sub-ogbn & 6860321 & 243561 & 0.012 & 0.009 \\
MB & 285423 & 52645 & 0.185 & 0.068 \\ \bottomrule
\end{tabular}
\begin{tablenotes}
\scriptsize
\item Vacant positions ("-") indicates the value does not exist. $C_i$ denotes the clustering coefficient of node $i$ and $Mean(C_i)$ denotes the mean of the clustering coefficients of all nodes. $\Delta_{Mean}$ = $Mean(C_i)$ after triangle augmentation strategies - $Mean(C_i)$ for the original data.
\end{tablenotes}
\end{threeparttable}

\end{table}

\subsection{\label{key-parameters} Key Hyper-parameters (for Q5)}

\begin{figure}
	\centering
	\subfigure[The Harmonic Coefficient $\lambda_U$]{
		\begin{minipage}[t]{0.3\linewidth}
		\centering
			\includegraphics[width=0.9\textwidth]{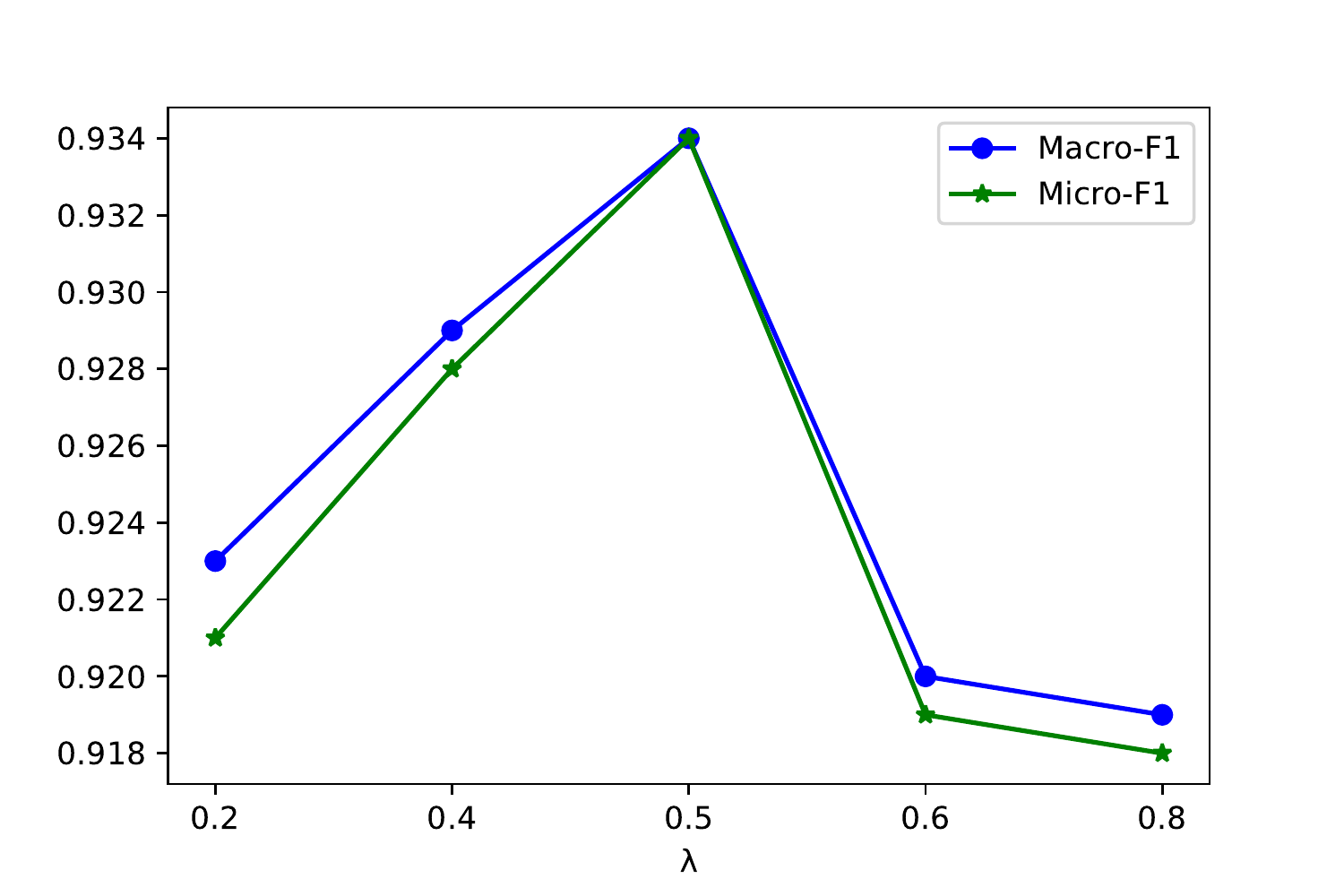}
		\end{minipage}
	} 
    	\subfigure[The Temperature $T$]{
    		\begin{minipage}[t]{0.3\linewidth}
    		\centering
   		 	\includegraphics[width=0.9\textwidth]{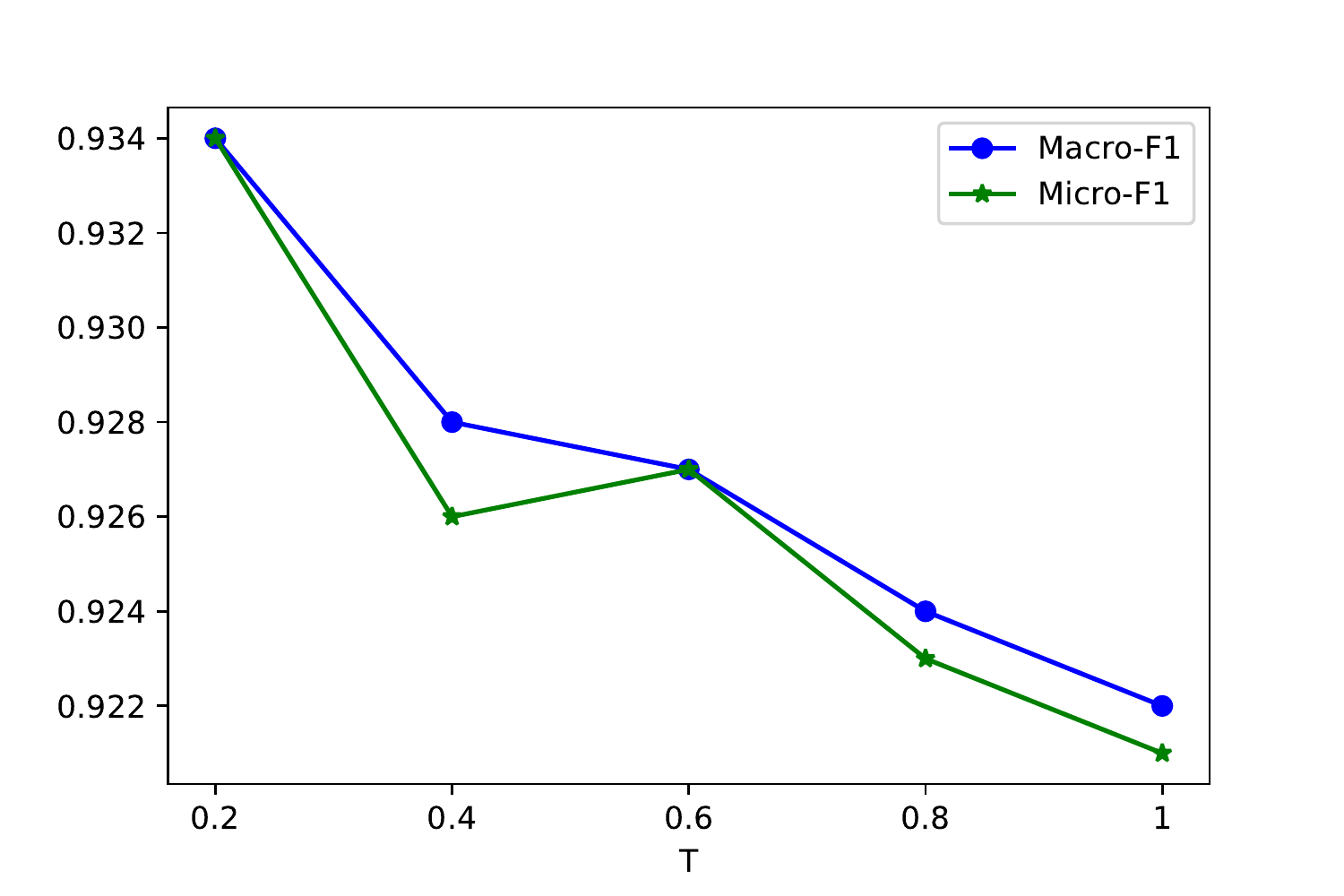}
    		\end{minipage}
    	} 
    	\subfigure[The Augmentation Ratio $K$]{
    		\begin{minipage}[t]{0.3\linewidth}
    		\centering
   		 	\includegraphics[width=0.9\textwidth]{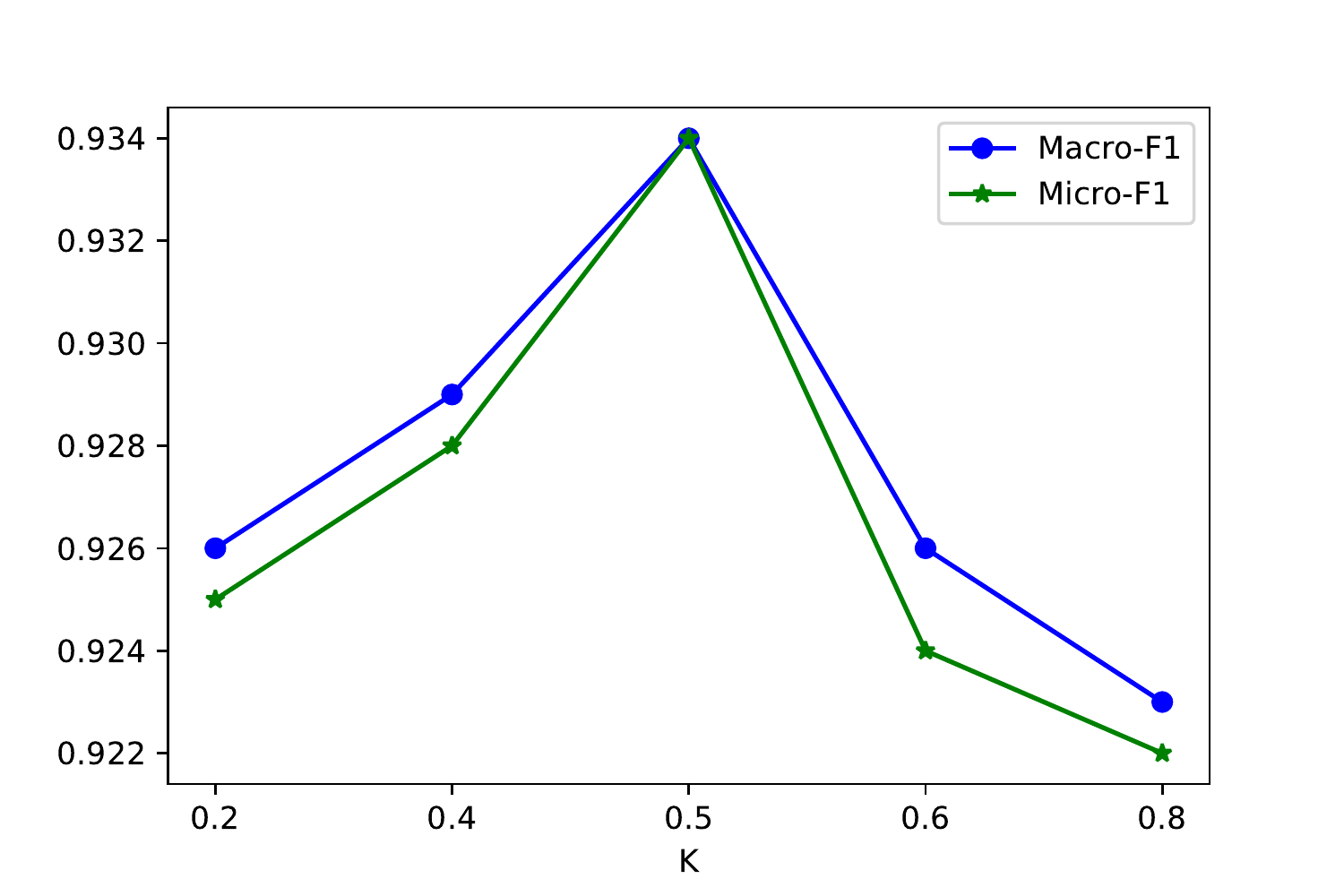}
    		\end{minipage}
    	}
	\caption{The Results with Different Values of The Key Parameters}
	\label{fig:kp}
\end{figure}

Experiments were performed with HG-MDA to discuss the effects of several key parameters in the model. The harmonic coefficient $\lambda_U$ represents the contribution ratio of the loss for unlabeled data to the final loss. The temperature $T$ is used in the loss for unlabeled data to adjust the relative magnitude between different augmentation results. Augmentation ratio $K$ is the sampling ratio that selects the number of nodes that need to be augmented. Figure \ref{fig:kp} shows the results obtained with different values of the key parameters.

When the harmonic coefficient $\lambda_U$ is between 0.4 and 0.6, the effect of the model decreases. The best result is obtained when $\lambda_U$ is about 0.5. It shows that both labeled supervision and unlabeled comparative learning have a certain positive impact on the model. Appropriately increasing the information of unlabeled data can help the model to train better. But when the effect of unlabeled loss is too large, the performance of the model will be degraded.

The smaller the value of $T$ is, the better the prediction result of the model is. It can be seen from the following formula that all augmented predicted values are multiplied by the reciprocal of $T$. The smaller T is, the classes with relatively large predicted relative probabilities will be sharply enlarged, which can make the network output more "confidence" forecast.

\[\bar{Z}^{'}_{uj} = \frac{\bar{Z}^{\frac{1}{T}}_{uj}}{\sum_{c=0}^{C-1}\bar{Z}^{\frac{1}{T}}_{uc}}(0\leq j \leq{C-1)}\]

When the augmentation ratio $K$ is also between 0.4 and 0.6, the model effect turns, and the effect is about 0.5. The more nodes that are augmented by data, the better. Adding a certain percentage of augmentation to the model can increase the connectivity between nodes and improve the negative curvature of the network, thereby enhancing the robustness of the model. However, when the augmentation ratio exceeds a certain threshold, the disturbance introduced in the model will mask the real information, which will have a negative impact on the model.

\section{Real-world Application scenario}

In the marketing pricing scenario of internet finance, according to historical data analysis, it is found that for people with strong financial needs, their behaviors such as visiting, clicking, and spending in financial products are very active, and they are sensitive to interests. Besides, the transactions and consumption of users with strong financial needs are similar. For example, among users who have a transfer relation with users with strong financial needs, the proportion of users with strong financial needs is also significantly higher. And there are a lot of relations in our scene, such as transfer relation, communication relation and friend relation. So, such a social networks can reflect the needs relation between users.

Our marketing target users are new customers with strong financial needs. However, for a user who has never used our products in history, his historical behavior data is lacked to judge whether he has strong financial needs, so we can only label users with strong financial needs through questionnaires. Relying on manual methods to mark new customers requires expensive time and labor costs, but get a little labeled data. Attempts to use a small amount of labeled data to train a supervised model are severely biased and the effect of making predictions on the remaining unknown data is unstable. Hence, unlabeled data is expected to use during model training through data augmentation. Additionally, Through a long-term data observation, the proportion of new customers who are currently unidentified but will be converted into strong money needs in the future is the same as the current known proportion of new customers with strong money needs, which means that the unlabeled data and the labeled data have the same distribution, which satisfies the conditions for semi-supervised learning. 

\begin{figure}
\centering
\includegraphics[width=0.5\textwidth]{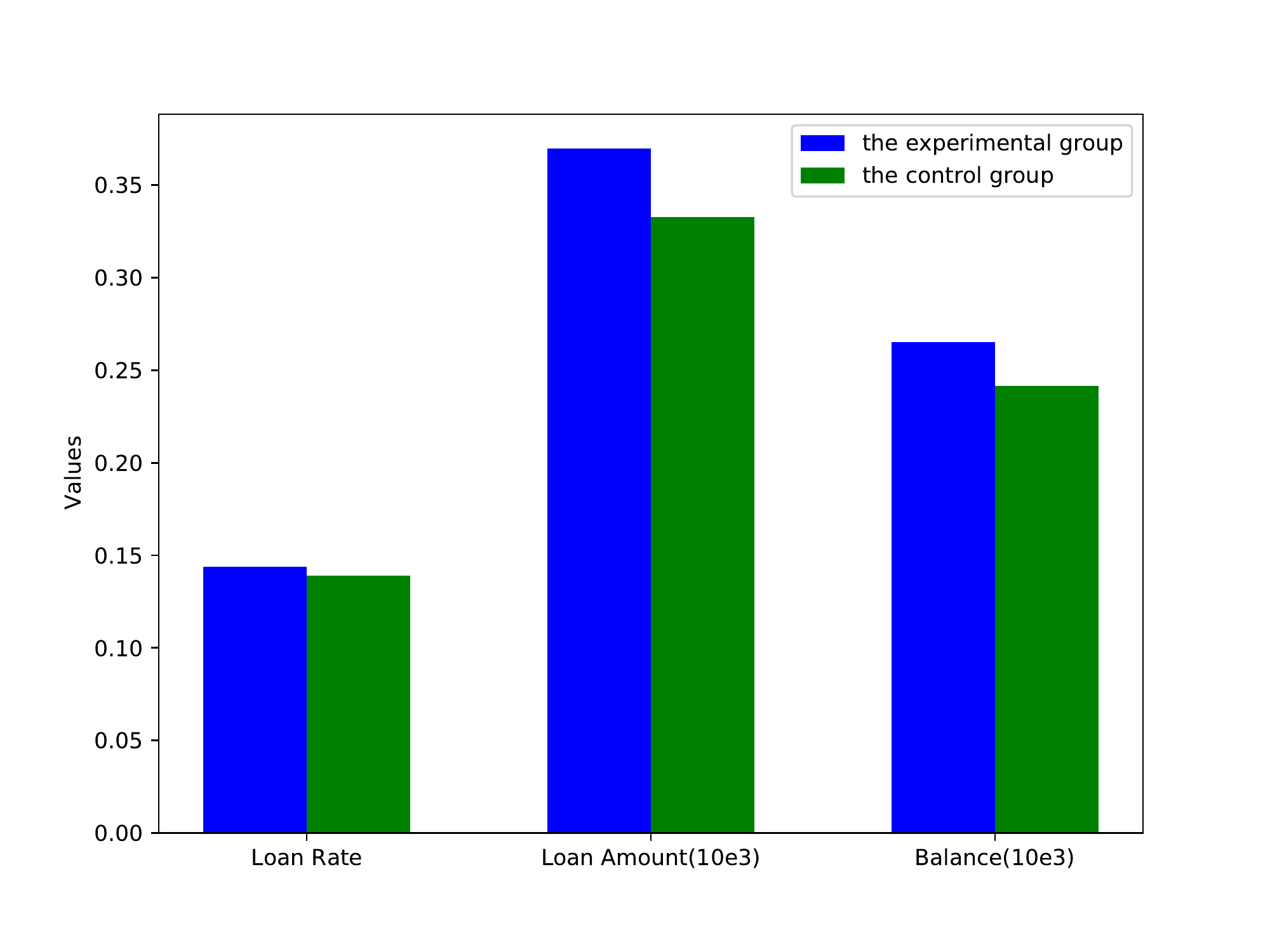}
\caption{\label{fig:ab_test}The Application Performance of HG-MDA in AB Experiment.  \textsl{For a nice display, the loan amount and balance by $10e3$ on the same scale as the loan rate. There are two homogeneous populations in the experimental group and the control group. In the experimental group, a certain proportion of users with strong financial needs are selected by the model for marketing, while in the control group, the same proportion of users is randomly selected.}}
\end{figure}

In summary, we apply HG-MDA to build a user identification model to identify new customers with strong financial needs, further improving the performance and generalization of the model. This model helps the business add 30\% of key users in the marketing scenario and can be reused in other scenarios. In the AB experiment shown as Figure \ref{fig:ab_test}, compared with the control group, the experimental group increased the loan rate by 3.6\%, loan amount increased by 11.1\%, and the balance increment increased by 9.8\%.

\section{Conclusion}

In this paper, we study the semi-supervised learning problem of heterogeneous graph by combining heterogeneous graph learning with DA strategies to enhance the generalization performance of the model. We propose a novel method named HG-MDA, that proposes three data augmentation strategies at node-level and topology-level to change the attributes of nodes and the triangular distribution in the graph structure according to the characteristics of heterogeneous graph. Considering that the information derived from different levels has different effects on model learning, feature exchange augment local information, while triangular augmentation augment structural information. We highlight that the triangle based edge adding and removing bring information gain compared to random edge adding and removing. And triangle is very stable structures in the network, which increases the stability of information propagation to a certain extent. In addition, in the process of DA, meta-relation-based attention is used to learn the information preference of the target node for different types of neighbor nodes and edges. The supervised loss and the consistency regularization loss are simultaneously minimized to fully utilize the unlabeled data. In order to avoid mutual interference between different augmentation strategies, the loss function is optimized by sharpening to combine their results effectively, which also contributes to the performance improvement.

Experimental results show that the proposed HG-MDA method achieves advanced performance in both node classification and node clustering tasks on three benchmark datasets ACM, DBLP, sub-ogbn and industrial dataset MB. In future work, we will further introduce other advanced DA strategies to improve our approach.

\bibliographystyle{ACM-Reference-Format}
\bibliography{bibfile}

\end{document}